\pdfoutput=1
\documentclass[11pt]{article}

\usepackage{EMNLP2022}

\usepackage{times}
\usepackage{latexsym}
\usepackage{algorithm}
\usepackage{stfloats}
\usepackage{subfigure}
\usepackage{tabularx}
\usepackage{array}
\usepackage{booktabs}
\usepackage{setspace}
\usepackage{url}
\usepackage{enumerate}
\usepackage{enumitem}
\usepackage{graphicx}
\usepackage{makecell}
\usepackage{multirow}
\usepackage{amsmath}
\usepackage{color}

\usepackage[T1]{fontenc}
\usepackage[utf8]{inputenc}

\usepackage{microtype}

\usepackage{inconsolata}

\title{FastClass: A Time-Efficient Approach to Weakly-Supervised Text Classification}

\author{Tingyu Xia$^{1,3}$, Yue Wang$^{2}$,
Yuan Tian$^{1,3,*}$,
Yi Chang$^{1,3,4,}\thanks{{\; Corresponding authors. }}$
	\\ 
	 \textsuperscript{\rm 1}School of Artificial Intelligence, Jilin University\\ 
     \textsuperscript{\rm 2}School of Information and Library Science, University of North Carolina at Chapel Hill\\
	 \textsuperscript{\rm 3}Key Laboratory of Symbolic Computation and Knowledge Engineering, Jilin University\\ 
	 \textsuperscript{\rm 4}International Center of Future Science, Jilin University	\\
	 xiaty21@mails.jlu.edu.cn, wangyue@unc.edu \\
	 yuantian@jlu.edu.cn, yichang@jlu.edu.cn\\
}

\begin{document}
\maketitle
\begin{abstract}
Weakly-supervised text classification aims to train a classifier  using only class descriptions and unlabeled data.
Recent research shows that keyword-driven methods can achieve state-of-the-art performance on various tasks. However, these methods not only rely on carefully-crafted class descriptions to obtain class-specific keywords but also require substantial amount of unlabeled data and takes a long time to train.
This paper proposes \textbf{FastClass}, an efficient weakly-supervised classification approach.
It uses dense text representation to retrieve class-relevant documents from external unlabeled corpus and selects an optimal subset to train a classifier. 
Compared to keyword-driven methods, our approach is less reliant on initial class descriptions as it no longer needs to expand each class description into a set of class-specific keywords.
Experiments on a wide range of classification tasks show that the proposed approach frequently outperforms keyword-driven models in terms of classification accuracy and often enjoys orders-of-magnitude faster training speed.

\end{abstract}

\section{Introduction}

% we propose a new method that achieve ideal scenario. Basic idea: leverage dense retrieval models to obtain weakly labeled documents from 1) test data; or 2) external corpus. 
% related to previous ideas: pseudo-relevance feedback (weak labels are obtained using top-ranked documents); entailment model (dense retriever is learned from entailment task data)
% experiments comparing our approach to ___ baselines show good performance (accuracy, speed)
% main contributions 

Text classification is one of the most used techniques in mining large-scale unstructured text. When sufficient labeled data are available, supervised classification techniques can achieve excellent performance. However, manually labeling example documents can be time-consuming and labor-intensive, a major burden when applying supervised text classification techniques in practice.

Recently, \emph{weakly-supervised text classification}~\cite{meng2018weakly, meng2019weakly, meng2020text, mekala2020contextualized,  shen2021taxoclass, wang2021x, zhang2021weakly} has been proposed to save labeling efforts. It refers to the ability for a machine learning model to start classifying documents by using only class descriptions and unlabeled data. 
Since weakly-supervised text classification methods do not rely on any labeled data, it can greatly reduce the workload of data labeling.
%Since any text classification task necessarily starts with a description for each class, class descriptions are naturally available from the very beginning.
Therefore, these methods are desirable in  real-world text mining applications. 

The current mainstream weakly-supervised text classification methods are keyword-driven~\cite{meng2018weakly, meng2019weakly, meng2020text, mekala2020contextualized,  shen2021taxoclass, wang2021x, zhang2021weakly}, where the users need to carefully choose initial class-relevant keywords for each class. Such keywords are expanded into a richer set of keywords, which are then used to assign pseudo-labels on unlabeled data. The critical step of these keyword-driven methods is the choice of initial class-relevant keywords and the subsequent keyword expansion strategy, which determines the quality of pseudo-labeled documents. 

%In the early research, ConWea~\cite{mekala2020contextualized} leveraged contextualized representation to disambiguate keywords. LoTClass~\cite{meng2020text} expanded the keywords through the MLM module of BERT.  X-Class~\cite{wang2021x} understood the semantics of keywords from the perspective of representation learning. ClassKG~\cite{zhang2021weakly} utilized keyword graph to obtain the correlation among different keywords, and transforming the task of assigning pseudo-labels for unlabeled documents into subgraphs annotating task. 

However,  
if the user's initial choice of  class descriptions is ambiguous or brief, then  the results of keyword expansion will be negatively impacted. %the performance of the model will be greatly reduced. 
Ideally, the user can provide initial class descriptions that are beneficial to keywords expansion. However, in reality, the quality of class descriptions is not guaranteed.
Early research often artificially modifies original class descriptions to improve the quality of the initial keywords~\cite{chang2008importance}.
In sentiment classification tasks, ``positive'' and ``negative'' class descriptions have to be replaced by ``good'' and ``bad'', which are easier for  keywords expansion~\cite{meng2020text}. 
%This results in a less robust model which works well in general classification tasks is less effective when facing low-quality datasets.

Previous weakly-supervised methods also face the common problem of fully supervised methods. That is, model performance depends on the amount of (unlabeled) data available.  Widely used classification datasets in research literature often have high-quality class descriptions and sufficient unlabeled data. However, weakly-supervised classification methods are most needed in the early stage of practical text mining tasks, where practitioners are not ready to invest vast amounts of efforts in data labeling or computing resources in model training. In this stage,  class descriptions can be crude and even unlabeled data can be scarce (e.g., classifying text related to an emerging event).

As we will show in our experiments, the increasingly sophisticated keyword-driven weakly-supervised methods require increasing amount of computing resources and  time in model training, and yet the resulting performance benefits can be minimal. This is concerning given the recent recommendation on Green AI~\cite{schwartz2020green} and Efficient NLP~\cite{arase2021efficient}.
Recent research found that Transformer-based textual entailment models can provide more competitive performance on dataless classification tasks~\cite{yin2019benchmarking, chu2020natcat} which is similar to weakly-surpervised text classification. In principle, such models only need to be trained once and can be applied to any classification tasks. However, it is not only difficult to adapt such models to a specific corpus, but also inefficient to run them at prediction time. That is, one has to run the entailment model $k$ times to classify one document into $k$ categories. The prediction speed slows down as more categories (larger $k$) are considered in a task.  

In this paper, we propose \textbf{FastClass}, a time-efficient weakly-supervised text classification model  which constructs a classification model by selecting a optimal subset of retrieved documents to solve the above problems.
\textbf{FastClass} has three steps: (1) extracting seed unlabeled documents with high semantic similarity to the original class descriptions; (2) using seed documents as queries to obtain pseudo-labeled documents from external corpus. (3) selecting an optimal subset of pseudo-labels  using the maximum entropy principle.

Our main contributions are as follows:
\begin{itemize}
\item We propose an efficient weakly-surpersived text classification method that selects a document subset returned by dense retrieval models as pseudo-labels for classifier training.
% (Section \ref{sec:method}). 
\item Compared to keyword-driven weakly-supervised methods, our method is less sensitive to class descriptions and does not heavily rely on high-quality category descriptions.
%still achieves good results with only low-quality raw class descriptions.
\item Extensive empirical experiments show that our method has higher accuracy and faster training speed in most cases, and the scale of task data does not have a significant impact on training time and model performance.
% (Section \ref{sec:exp}).
\end{itemize}

% Then the question is: how to get such a classification model that can not only entailment textual knowledge, but also can flexibly adapt to downstream testing tasks? In this article, we extract the knowledge of entailment model and teach it to a classification model. The refining process is: use SBERT\cite{reimers-2019-sentence-bert}, which is trained with entailment data, and extract texts based on the semantic similarity of class labels. These data can come from test data, or from a large amount of external data. Then these data can be used as weakly labeled data to train a classification model. Experiments have proved that the classification model obtained in this way not only solves the two shortcomings of the above-mentioned entailment models, but also the accuracy is higher than the entailment models most of the time.

\section{Related Work}
We discuss related work from three perspectives:  zero-shot text classification, dataless text classification, and weakly-supervised text classification.

\subsection{Zero-Shot Text Classification}
Zero-shot text classification divides classes into seen classes with annotated data and unseen classes without any labeled data. 
Since no labeled data is available for unseen
classes during training, the general idea of zero-shot learning is to transfer knowledge from seen classes to unseen classes.

The mainstream zero-shot text classification method is to exploit semantic knowledge to generally infer the features of unseen classes using patterns learned from seen classes. Three main types of semantic knowledge have been employed in general zero-shot scenarios, including semantic attributes and properties of classes~\cite{liu2019reconstructing, xia2018zero, pushp2017train}, correlations among classes~\cite{rios2018few, zhang2019integrating}, semantic word embeddings which capture implicit relationships between classes and documents~\cite{nam2016all}. Besides, \citeauthor{ye2020zero} explored reinforcement learning framework to tackle zero-shot task.

\subsection{Dataless Text Classification}
Dataless text classification~\cite{chang2008importance} aims to classify text using a given set of class descriptions and no labeled data for training a model. These methods have two broad categories: classification-based~\cite{song2014dataless, yin2019benchmarking,  chu2020natcat} and clustering-based~\cite{li2018dataless, li2018pseudo}. 

In previous works, dataless text classification also has many slightly different setups. For example, in zero-shot text classification, \citeauthor{yin2019benchmarking} proposed  ``label-fully-unseen'' setting which directly computes document-label relatedness with a sentence-pair BERT model. The model is trained with large-scale  texts naturally tagged with category information, such as Wikipedia. NATCAT~\cite{chu2020natcat} combines various publicly available online corpora that come with natural categories, and trains a BERT~\cite{devlin2018bert} or RoBERTa~\cite{ liu2019roberta}model to discriminate correct versus incorrect categories for a given document. These methods create pseudo-labeled data from external resources to train a universal textual entailment model that can be applied to a wide spectrum of classification tasks. 

\subsection{Weakly-superivised Text Classification}
It is easy to confuse ``dataless text classification'' with ``zero-shot text classification''~\cite{wang2019survey,ye2020zero} and ``weakly-supervised text classification''~\cite{meng2019weakly,meng2020text}.
Zero-shot text classification may still provide labeled data for part of the categories (label-partially-seen \cite{yin2019benchmarking}), while
dataless text classification can be applied to any text classification task with a given set of label descriptions by building a general text classifier. 
Weakly-supervised text classification assumes a large amount of unlabeled data and label descriptions are available for training.

Most previous works on weakly-supervised text classification are keyword-driven methods. These methods generate pseudo-labeled data by counting class-relevant keywords in unlabeled documents and then train a supervised model. Among them, WeSTClass~\cite{meng2018weakly}  utilizes a self-training module which bootstraps on unlabeled data for model refinement. WeSHClass~\cite{meng2019weakly} extends WeSTClass to hierarchical labels. LOTClass~\cite{meng2020text} uses label names as initial keywords and augments the keywords with BERT's MLM module. ConWea~\cite{mekala2020contextualized} disambiguates polysemy keywords based on contextualized corpus. 
X-Class~\cite{wang2021x} constructs pseudo-labeled document by estimating class-oriented document representation and document clustering.
ClassKG~\cite{zhang2021weakly} improves the quality of pseudo-labels by applying GNN to a keyword graph to exploit keyword correlation.

\section{Proposed Methods}
\label{sec:method}

In this section, we describe our proposed method \textbf{FastClass} for weakly-supervised text classification. We formulate the problem as follows. We are given a set of class descriptions $D = \{d_1, \cdots, d_j, \cdots, d_k\}$, each is a  piece of short text (one or more words) describing a semantic class $j$ in the label space $Y = \{1, \cdots, k\}$. We are given a set of unlabeled documents $X$ in the task domain. As a natural scenario in practice, we also have access to vast amounts of external unlabeled documents $U$, $|U| >> |X|$. These external documents may come from Wikipedia, news corpora, and online social media, which may or may not share the same domain as the classification task in question.
Our goal is to correctly assign label(s) from $Y$ to (a subset of) unlabeled documents in $U$ as pseudo-labeled training data.

\begin{figure*}[htbp]
\centering
\includegraphics[width=\textwidth]{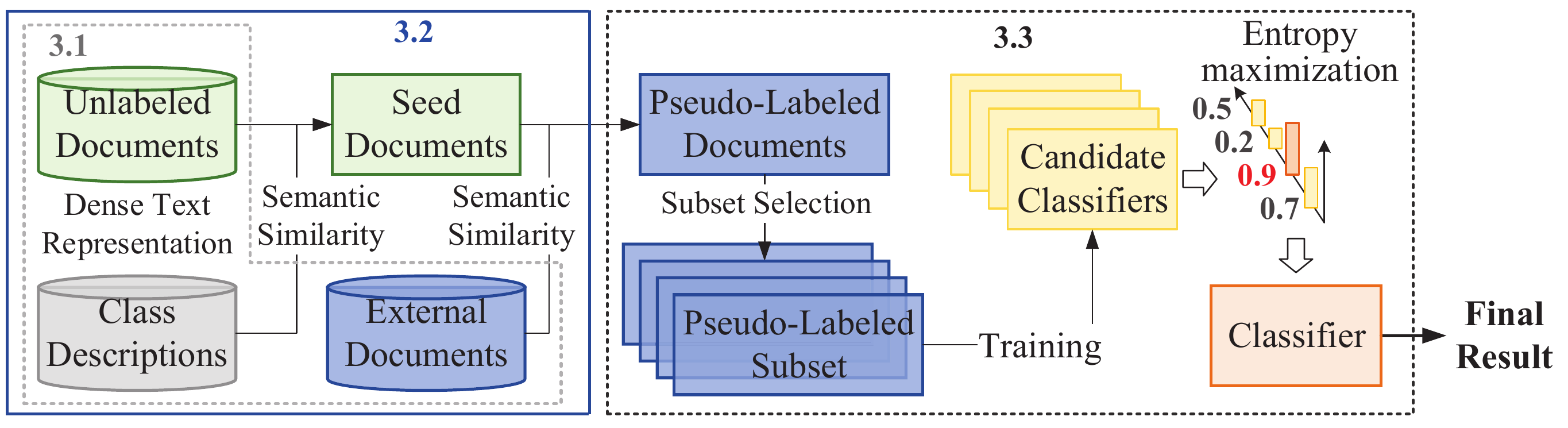}
\caption{An overview of the \textbf{FastClass} approach. We first get dense representations of unlabeled documents, class descriptions and external documents (Section \ref{sec:textemb}). Then we obtain task-specific seed documents from unlabeled documents and use them as queries to retrieve pseudo-labeled documents from external documents (Section \ref{sec:retrieve}). Lastly, we select the optimal subset of these pseudo-labeled documents based on entropy maximization strategy (Section \ref{sec:subset}). Different colors represent different origins of data.}
\label{fig:model_struc}
\end{figure*}

At a high level, our proposed method uses class descriptions in $D$ to obtain task-specific documents that are highly similar to class descriptions from $X$ and  uses these task-specific documents as queries to retrieve pseudo-labeled documents from external unlabeled data $U$. Then we select the optimal subset of these pseudo-labeled documents to train a classifier. The general \textbf{FastClass} framework is presented in Figure~\ref{fig:model_struc}.
Below we describe our method in detail.

% how to construct a pseudo training set to train a classification model for dataless text classification. We provide three pseudo training set construction methods: self-data construction, external data construction and internal and external data joint construction.

% We will introduce them in detail in further sections.

% \begin{figure}[h]
%   \centering
%   \includegraphics[width=\linewidth]{LaTeX/figure/figure1.pdf}
%   \caption{ Overview of our Two-Stage framework for dataless text classification.}
%   \end{figure}

\subsection{Dense Text Representation and Indexing}
\label{sec:textemb}
As a preparation step, we use a sentence representation model to convert all texts (class descriptions, task-specific unlabeled documents, and external unlabeled documents) into dense vectors in a semantic space. In principle, any dense text representation techniques can be used. We choose to use Sentence-BERT (SBERT) \cite{reimers-2019-sentence-bert} as it is proven to deliver good performance in various sentence-pair modeling and information retrieval tasks \cite{thakur2021beir}. 

Once these texts are converted into dense vectors, we build approximate nearest neighbor (ANN) indices for task-specific unlabeled documents and external documents to enable fast document retrieval. In principle, any ANN search techniques can be used. We choose to use FAISS~\cite{JDH17} for efficient similarity search with cosine similarity as the vector similarity metric. We also tested other metrics such as Euclidean distance but found negligible performance difference. 

%As SBERT is trained on a wide range of semantic similarity tasks (including textual entailment), the resulting document vectors inherit the knowledge from these tasks. Cosine similarity $\cos(x_1, x_2)$ between documents $x_1$ and $x_2$  approximates the probability that $x_1$ entails $x_2$ (or vice versa). In this sense, our method implicitly leverages the same type of knowledge of entailment-based models in a more efficiently computable manner.

\subsection{Class-Relevant Document Retrieval}
\label{sec:retrieve}
The first step of our method is to retrieve a pool of potentially relevant documents for each class, a subset of which are pseudo-labeled in the next step. %We propose two variants for this step.

We retrieve pseudo-labeled documents from external data with a task-specific focus. The idea is to enrich a class description with task-specific data before  retrieving from external documents.
For each class $j$, we first obtain a ``seed set'' of documents $S_j$.
We use each class description as a search query to retrieve documents from task-specific unlabeled data. For class $j \in Y$, we rank documents in the unlabeled data $X$ by their semantic similarity to the class description $d_j$ and take the most similar $c$ documents $S_j = \{x_i\}_{i=1}^{c}$. Here, semantic similarity is computed using the vectors produced in Section \ref{sec:textemb}.
%using the same approach as {\sc Claret}$_{\text{task}}$ by fixing $n_1 = .1 \times |X|/k$.  
Then we use $S_j$ to further retrieve external documents by treating each $x \in S_j$ as a query to retrieve its $n$ nearest neighbors $\Gamma(x)$ from external data. However, these documents may be close to a seed document because they share words unrelated to the theme of the class. To filter such noise, we preserve documents that appear in at least two seed documents' nearest neighborhoods. This gives class-relevant documents for class $j$: $R_j = \{e| \exists x_1, x_2 \in S_j, e \in \Gamma(x_1) \wedge e \in \Gamma(x_2) \}$.  The  hope is that $R_j$ contains external documents that are  semantically relevant and stylistically similar to task-specific unlabeled data. 

\begin{figure*}[h]
\centering
\includegraphics[width=0.9\textwidth]{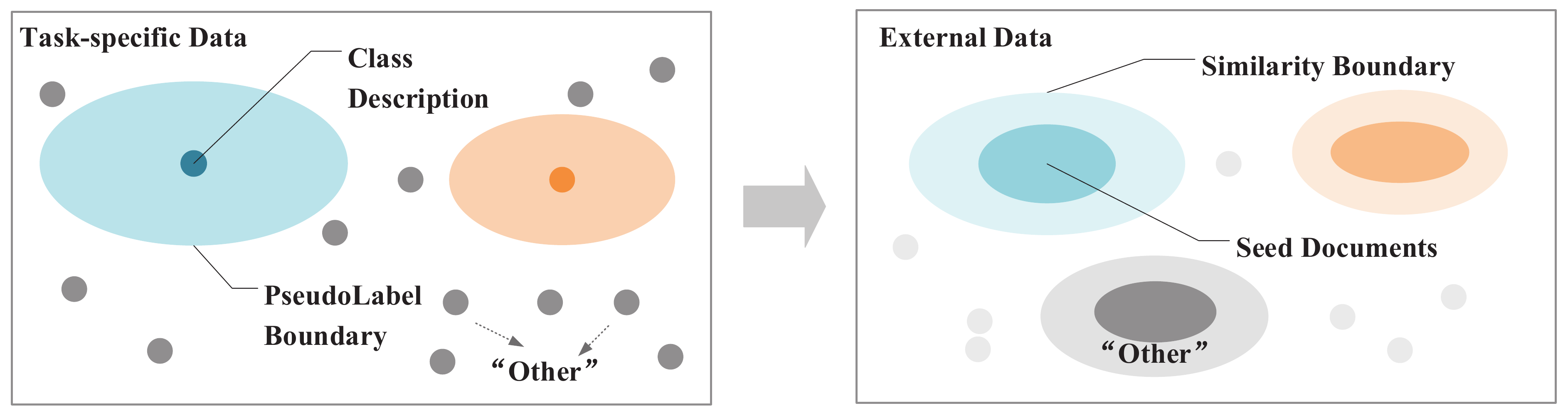}
\caption{Handling the \textit{Other} class. }
\label{fig:other}
\end{figure*}

\subsection{Pseudo-Labeled Subset Selection}
\label{sec:subset}

% \begin{algorithm}[!t]  
% 	\caption{FastClass}  \label{alg:fastclass_algorithm}
% 	\begin{algorithmic}[1]
% 		\Require $D$, $X$, $U$
% 		\State $c = .1 \times |X|/k$
% 		\For {each $d_{j} \in D $}
%     		\State 	$X_{j} = CosineSimilarity(X, d_{j})$
%     		\State $S_j = \{x_i\}_{i=1}^{c}, (x \in X_{j})$
%     		\State  $\Gamma(x) = CosineSimilarity(x, U) ,(x \in S_{j})$
%     		\State  $R_j = \{e| \exists x_1, x_2 \in S_j, e \in \Gamma(x_1) \wedge e \in \Gamma(x_2) \}$ 
%     	\EndFor
% 		\While {$E^{r}_{cur} - E^{r}_{mem}$ is not empty}
%     		\For {each $e^{r}_{i} \in E^{r}_{cur} - E^{r}_{mem}$}
%         		\State  $V_i$ = $\mathbf{GetLinkedEntNodes}$($\mathcal{G}, e^{r}_{i}$)
%         		\State add rule $e^{r}_{i}$ and nodes $V_i$ to $\mathcal{G}^Q$
%         		%\For {each $v_{j} \in V_i$}
%             	%	\State add edge between $v^{r}_{i}$ and $v_{j}$ to $\mathcal{G}^Q$
%         		%\EndFor
%     		\EndFor
%     		\State expand $E^{r}_{mem}$ with $E^{r}_{cur}$
% 		    \State assign $V_{cur}$ with nodes in $\mathcal{G}^Q$
%     		\State  $E^{r}_{cur}$ = $ \mathbf{GetLinkedRuleEdges}$($\mathcal{G},V_{cur}$)
% 		\EndWhile
% 		\State \Return InferGraph $\mathcal{G}^Q$
% 	\end{algorithmic}
% \end{algorithm} 

\textbf{Subset diversification}. 
We select a subset $L_j \subset R_j$ of size $m$ to be pseudo-labeled as class $j$. The motivation is that documents retrieved from external data sources may contain (near-)duplicates. For example, many news outlets may cover the same story. Duplicated documents may lead to overfitting as they give too much emphasis on a few documents and reduce the overall diversity of pseudo-labeled training data. Indeed, previous works have shown that diverse training data improves learning performance \cite{wei2015submodularity}. Here we apply facility location function  to quantify the diversity of a subset. The facility location function of  any subset $L_j \subset R_j$ is defined as 
\begin{align}
g(L_j) = \sum_{x \in R_j} \max_{e \in L_j} s(x,e) \ .
\end{align}
Here $s(\cdot, \cdot)$ is the cosine similarity between two dense document vectors. Intuitively, $g(L_j)$ computes the total cost for every element $x \in R_j$ to be ``covered'' by the most similar element $e \in L_j$. In our context, this translates into how well the subset $L_j$ preserves the content of the larger set $R_j$. Although finding the optimal subset $L_j$ that maximizes the submodular function $g(L_j)$ is NP-hard, a greedy algorithm gives an approximately optimal solution  \cite{nemhauser1978analysis}. The algorithm sequentially adds the next element $x$ to $L_j$ with the maximum marginal gain $g(L_j\cup\{x\}) - g(L_j)$, until $L_j$ reaches the desired size $m$. 

A challenging problem remains: how many documents to retrieve and assign pseudo-labels (namely, how to set $n$ and $m$)? More generally, what is the optimal subset of retrieved documents that, if pseudo-labeled, will train a good classifier? Note that we cannot tune subset selection procedures on labeled data as such data is unavailable in a weakly-supervised setting! To address this problem, we propose a novel \textit{unsupervised} subset selection procedure as follows.

\textbf{Entropy maximization}. We now determine the subset selection parameters $\theta = \{n, m\}$.
$\theta$ determines the pseudo-labeled set $L_j$ for class $j$, which determines the full pseudo-labeled set $\cup_{j=1}^k L_j$, which in turn trains a classifier $f: X \rightarrow Y$. Below we use $f_\theta$ to emphasize that $f$ depends on $\theta$. Once trained, $f_\theta$ induces a distribution over the label space $Y$  when applied to the task-specific unlabeled data $X$: $\forall y\in Y$, 
\begin{align}
    p(y|X,f_\theta) = \frac{\sum_{x\in X} \mathbf{1}\{f_\theta(x)=y\}}{|X|} \ .
\end{align}

According to the \emph{maximum entropy} principle \cite{jaynes1957information}, the distribution with maximum entropy shall be preferred since we have \textit{no} labeled data as evidence to prefer any other distribution. 
Following this principle, we seek for $\theta$ that maximizes the $f_\theta$-induced classification entropy:
\begin{align}
    H(\theta) = \sum_{y\in Y} -p(y|X,f_\theta) \log p(y|X,f_\theta) \ .
\end{align}
% $p(y|X,f_\theta) = \sum_{x\in X} \mathbf{1}\{f_\theta(x)=y\}/|X|$
Empirically, $H(\theta)$ correlates well (but not perfectly) with true performance of $f_\theta$ on labeled data even though it is an unsupervised metric  (Appendix \ref{sec:ent-acc-relation}), a phenomenon first observed in \cite{baram2004online}. As $H(\theta)$ is non-differentiable with respect to  $\theta$, we resort to grid search. It is  sufficient to use a coarse grid to find sensible $\theta$ values (Section \ref{sec:compared}).

\subsection{Handling the \textit{Other} Class}
\label{sec:other}
% previous methods have different strategies to deal with this situation
% problem formulation: given pool of documents, selected documents from named classes; want: "other" documents.
% method: find those documents that are farthest from "named" class documents [use a figure to illustrate]
% for a and c above: select "other" documents from S1 
% for b above: select "other" documents from S2
In some classification tasks, we have clearly defined categories and an \textit{Other} category, such as an ``other topic'' category in topic classification or a ``no emotion'' category in emotion classification. We call clearly defined (non-\textit{Other}) categories \textit{named classes}. 
Using ``other topic'' or ``no emotion'' literally as the search query to retrieve pseudo-labeled documents is problematic because the \textit{Other} class is to be interpreted with respect to named classes. 

The general idea is to pseudo-label documents that are  far from any named class as the \textit{Other}  class.  Without loss of generality, let the named classes be numbered from $1$ to $k-1$ and the \textit{Other} class be class $k$.

We first select \textit{Other} documents $L_k$ from task-specific unlabeled data $O = X \backslash \cup_{j=1}^{k-1} L_j$.  Our goal is to find a subset $L_k \subset O$ with size $c$ that is farthest from the descriptions of all named classes $D\backslash \{d_k\}$. We seek for the subset that \textit{minimizes} the following function:
\begin{align}
    h(L_k) = \sum_{x\in L_k}  \max_{ e \in D\backslash \{d_k\} } s(x,e) \ . \label{eq:other-fun}
\end{align}
This function is modular and can be efficiently minimized by selecting $c$ documents that have smallest $\max_{j=1}^{k-1} s(d_j, x)$ values from $O$ (Figure \ref{fig:other} left). 
This turns \textit{Other} into another named class. 
We then retrieve and select pseudo-labels from external data using the same procedure described in Sections \ref{sec:retrieve} and \ref{sec:subset}
(Figure \ref{fig:other} right).

%For {\sc Claret}$_{\text{external}}$, we first retrieve external data   that are far from all named class descriptions but still relevant to the task: $O = \cup_{j=1}^{k-1} \{x|x\in U, 0 < s(x,d_j) < 0.1\}$. We then select $R_k \subset O$ with size $n_1$ by optimizing $h(R_k)$ (Eq. \eqref{eq:other-fun}), and then use the same diversity and entropy maximization procedure in Section \ref{sec:subset} to select $m$ documents in $R_k$ and pseudo-label as \textit{Other} (Figure \ref{fig:other}b). 

\section{Experimental Evaluation}
\label{sec:exp}

\subsection{External Document Repository}
To cover various task domains, we combine five large-scale datasets as the external document repository. These datasets are freely available and frequently used in previous works as external resources. We keep these documents short (e.g. titles) as SBERT is well-trained on sentence pairs. We build a single index for all the external documents.

\textbf{Microsoft News Dataset (MIND)}~\cite{wu2020mind} is collected from anonymized behavior logs of Microsoft News website. 
\textbf{Multi-Domain Sentiment Dataset (MDSD)}~\cite{blitzer2007boom} contains product reviews for many product categories in Amazon. 
\textbf{Wikipedia-500K}~\cite{Bhatia16} has over a million curator-generated category labels and each article often has more than one relevant labels. We select the first sentence of each article.
\textbf{RealNews}~\cite{zellers2019grover} is a large news corpus from Common Crawl. We randomly sample 2M titles from these 32M news.
\textbf{S2ORC}~\cite{lo-wang-2020-s2orc} is a general corpus of  scientific literature. We randomly select 100k papers from all 20 research fields and extract their titles. The statistics of datasets are shown in Table \ref{tab:repository}.

\begin{table}[htbp]
  \setlength{\abovecaptionskip}{2.5pt}
  \centering

   \setlength{\belowcaptionskip}{3pt}
  \renewcommand\arraystretch{1.1}
    \renewcommand\tabcolsep{4pt}
  \label{tab:external-datasets}
    \begin{tabular}{lm{2cm}<{\raggedleft}m{2cm}<{\raggedleft}}
    %{m{2cm}|p{25pt}<{\raggedleft}p{25pt}<{\centering}p{35pt}<{\centering}p{35pt}<{\centering}}
      \toprule
      Dataset  &\#Docs &\#Words/doc\\
      \hline
      MIND &285,875 &10.7 \\
      MDSD &821,249  &137.5  \\
      Wikipedia   &1,779,881 &22.9 \\
      RealNews &2,000,000  &9.6\\
      S2ORC &2,000,000 &10.9\\
      \toprule
    \end{tabular}
  \caption{Statistics of external document datasets.}
  \label{tab:repository}
\end{table}

\subsection{Evaluation Datasets}
We choose 9 text classification tasks in our experiments. Note that we only use the original class descriptions (see Appendix \ref{app:class-name}). 

% None of the descriptions are manually expanded to include synonyms or paraphrases.
\textbf{Yahoo}~\cite{zhang2015character} consists of 10 categories of questions in online forums.
\textbf{20Newsgroup}~\cite{lang1995newsweeder} is a collection of 20 topic newsgroup documents.
\textbf{AGnews}~\cite{zhang2015character} contains 4 topical categories of news tities.
\textbf{DBPedia}~\cite{lehmann2015dbpedia} contains titles, descriptions, and associated categories  from DBpedia. 
\textbf{Yelp}~\cite{zhang2015character} is for sentiment analysis in Yelp reviews. 
\textbf{Emotion}~\cite{oberlander2018analysis} was constructed by combining  multiple public datasets where documents have emotion labels. 
\textbf{Amazon}~\cite{zhang2015character} is a binary sentiment classification dataset.
\textbf{SST}~\cite{socher2013recursive} is a corpus extracted from movie reviews. \textbf{Situation}~\cite{zhang2015character} is a event-type classification dataset originally designed for low-resource situation detection. 
The statistics of datasets are shown in Table \ref{tab:datasets}.

\begin{table}[h]
  \centering
  \renewcommand\arraystretch{1.1}
  \renewcommand\tabcolsep{5pt}
  \label{tab:task-datasets}
    \begin{tabular}{m{1.8cm}p{35pt}<{\centering}p{35pt}<{\centering}p{35pt}<{\centering}}
      \toprule
      Dataset  &\#Classes  &\#train &\#test\\
      \midrule
      Yahoo     &10  &1300k   &100k \\
      AGnews    &4   &120k    &7,600\\
      20News    &20  &11,015  &7,318\\
      DBPedia   &14  &560k    &70k\\
      Yelp      &2   &560k    &38k\\
      Emotion   &10  &34,667  &16k\\
      Amazon    &2   &3600k   &400k\\
      SST-B     &2   &67,349  &1,821\\
      Situation &12  &4,921   &3,495\\
      \toprule
    \end{tabular}
    %}
  \vspace{-5pt}
  \caption{Statistics of evaluation datasets.}
  \label{tab:datasets}
\end{table}

\begin{table*}[t]
\renewcommand\arraystretch{1} 
\renewcommand\tabcolsep{5pt}

	\centering
	\begin{tabular}{l|ccccccccc}
	\toprule
	Method &Yahoo &AGnews &20News &DBPedia &Yelp &Emotion &Amazon &SST  &Situation \\
	\midrule
    %\multicolumn{10}{l}{\textbf{Dataless}}\\

	0SHOT-TC & 34.8 & 53.8 & 22.2 & 53.8 & 73.4 &\underline{21.7} & 76.0 & \underline{71.7} &16.2\\
	
	NATCAT &48.6  & 74.9 & 44.8 &85.3  & 50.1 &10.7 &50.8  & 50.5 &27.4 \\
	\midrule
	%\multicolumn{10}{l}{\textbf{Weakly-supervised}}\\

	LOTClass & 53.3 & 83.1 &/ & 90.9 &/ &/  &51.2 &49.6 &/ \\
	
	X-Class & 47.7 & \underline{85.7} &39.2 &91.5 &63.2  &13.1 &84.2 &48.3  &\textbf{41.1}  \\

	ClassKG & \underline{60.8} & \textbf{87.9} &\underline{49.4} & \textbf{98.3} & \textbf{92.5} 
	&/  & \underline{88.0} &67.0 &/ \\
   
    %\multicolumn{10}{l}{\textbf{\textsc{Claret}}}\\
    
    %External &57.3	&72.7	&51.7	&84.9 &87.9	&27.6	&89.5	&80.1	&30.5\\

    FastClass &\textbf{61.9}	&85.1	& \textbf{55.4}	&\underline{92.4}	&\underline{87.9}	 & \textbf{28.0} &\textbf{89.1}	& \textbf{84.5}	&\underline{39.2}\\ 
    
    \midrule
    \multicolumn{10}{l}{\textbf{\textsc{Supervised}}}\\
    RoBERTa &75.1	&95.4	&73.5	&99.3	&97.5	&37.8	&97.4	&95.8	&58.4\\
    
	\bottomrule
	\end{tabular}
\caption{Weakly-supervised text classification performance on nine datasets (\%). The metrics are label-weighted F1 for Emotion and Situation and accuracy for other tasks.  Mentioned that ``/'' indicates the current model can not obtain classification result of the current dataset. The best performance in each column is bold and second-best performance is underlined.} 
\label{tab:perf}
\end{table*}
\begin{table*}[h]
\renewcommand\arraystretch{1} 
\renewcommand\tabcolsep{3.8pt}
	\centering
	\begin{tabular}{l|ccccccccc}
	\toprule
	Method &Yahoo &AGnews &20News &DBPedia &Yelp &Emotion &Amazon &SST  &Situation \\
	\midrule

	LOTClass &15h$\downarrow$ &29m$\downarrow$ &/ &5.5h$\downarrow$ &/ &/  &15h$\downarrow$ &13.5m$\downarrow$ &/ \\
	
	X-Class &32.5h$\downarrow$ &75m$\uparrow$ &0.5h$\downarrow$ &8.3h$\downarrow$ &14.3h$\downarrow$  &24m$\downarrow$ &62h$\downarrow$ &35m$\downarrow$ &8m$\uparrow$  \\

	ClassKG &41.7h$\downarrow$ &90m$\uparrow$ &8.5h$\downarrow$ &13.7h$\uparrow$ &26.5h$\uparrow$ &/  &133.5h$\downarrow$ &54m$\downarrow$ &/ \\

    FastClass &1.7h	&22m	&1h	&1.5h	&40m	 &44m &3.5h	&12m	&33m\\ 
    
	\bottomrule
	\end{tabular}
\caption{Comparison of training time for weakly supervised methods. The arrow on the right shows the performance of the current model compared to FastClass. ``h'': hours; ``m'': minutes.  ``/'' has the same meaning as in Table \ref{tab:perf}.}
\label{tab:efficiency}
\end{table*}	

\subsection{Compared Methods}
\label{sec:compared}
We include two state-of-the-art methods for dataless text classification and three weakly-supervised text classification methods:

\textbf{Dataless text classification}. \textbf{\textit{Label-fully-unseen} 0SHOT-TC}~\cite{yin2019benchmarking} pushes ``zero-shot learning'' to the extreme -- no annotated data for any labels. It aims to classify documents without seeing any task-specific training data.  \textbf{NATCAT}~\cite{chu2020natcat} proposed to use large-scale, naturally annotated data to train robust entailment-based text classification models. These two methods both use readily available resources to train textual entailment models that can robustly handle a wide range of text classification tasks. 

\textbf{Weakly-supervised text classification}.
\textbf{LOTClass}~\cite{meng2020text} utilizes pre-trained LM to collect category-indicative words and generalizes the model via document level self-training on abundant unlabeled data.
\textbf{X-Class}~\cite{wang2021x} estimates class-oriented document representations based on pre-trained language models from a representation learning perspective, and then selects high-confidence clustered examples to form a pseudo-training set.
\textbf{ClassKG}~\cite{zhang2021weakly} designs a new pretext task based on the keyword graph to learn better representations of keyword subgraphs, with which the accuracy of pseudo-label
generation is improved, and thus improves classification performance.

\subsection{Experimental Settings}
Below, we summarize the implementation details that are key for reproducing results. 

We use ``paraphrase-MiniLM-L6-v2'' as the base model for SBERT to obtain the sentence embeddings and the dimension of embedding vectors is $384$. FAISS is used to retrieve external documents which works with inner product to compute cosine similarity. The number of clusters is set to $512$ and $3$ clusters are explored at search time. 
We implemented facility location subset selection using the Apricot library~\cite{schreiber2020apricot}, which provides cosine as a similarity measure and a lazy greedy optimizer as a solver. 

In our pilot study, we experimented with three approaches (Annoy, HNSWlib and FAISS) to building the approximate nearest neighbor index in the initial experimental phase. There was no substantial difference in their performance, so we chose FAISS which enjoys the fastest speed.

When retrieving seed documents, we set $c = .1 \times |X|/k$ to avoid too few pseudo-labels or too many inaccurate pseudo-labels per class. As a weakly supervised method, FastClass can directly operate on unlabeled  documents without distinguishing  training vs. test sets provided by a task. Therefore, we simply retrieve seed documents from unlabeled (i.e., test) documents to select an appropriate number of seeds, which adapt to the data size of the documents to be classified. 

To select pseudo-labeled subsets that have maximum classification entropy, we searched parameters $\theta$ on the grids $n = \{100, 200, 300\}$ and $m = \{300, 500, 800\}$. 
The subset-induced RoBERTa-base (110M parameters)\cite{liu2019roberta} classifier that achieved the maximum entropy was used. The optimizer is  AdamW~\cite{loshchilov2017decoupled}, learning rate is  $2e^{-5}$, training batch size is $32$ and the number of training epochs is $4$. 

The training and evaluation of all methods are performed on a NVIDIA GeForce RTX 3090 GPU with 24GB memory and an Intel Xeon Gold 6330 CPU with 14 cores and 80GB memory.

\section{Results}
In this section, we evaluate our proposed method and compare it with baseline models for dataless and weakly-supervised text classification. The comparison is not only in terms classification accuracy, but also model efficiency. 
\subsection{Performance Across Datasets}
Table \ref{tab:perf} summarizes classification performance of baseline methods and our pseudo-labeling method FastClass combined with RoBERTa classifier. We also show classification performance of FastClass based on BERT classifier in Appendix Table \ref{tab:bert_result}. For dataless methods, we report results of 0SHOT-TC based on the official pre-trained model used MNLI as training resource and our pre-trained NATCAT based on RoBERTa using Wikipedia as training resource. For weakly-supervised methods,  we report results of our reproduction of LOTClass, X-Class, and ClassKG.
All baseline results are obtained by ourselves based on the official codes. Besides, we have stored our implementation as open source code in our Github repository\footnote{\url{https://github.com/xiatingyu/FastClass}}.

We followed the metrics used in \cite{yin2019benchmarking} and chose label-weighted F1 as the metrics for unbalanced datasets Emotion and Situation. For other balanced datasets, we used accuracy.
Note that our method works on a weakly supervised setting, where the data has only texts but no labels. Our approach to generating suitable pseudo-labels for unlabeled texts is deterministic and does not involve randomness. Performing $k$-fold cross validation over pseudo-labels does not fit in our approach because classifier training is not the final step, but an inner-loop step of the entropy maximization procedure. So statistical significance tests are not applied in this work.

%For 0SHOT-TC, since the authors only provided the pre-trained BERT model, so in Table \ref{tab:perf}, we reported the results obtained by our own implementation of RoBERTa model using MNLI data.

%For single label classification tasks we report accuracy scores. For multi-label classification tasks we report label ranking average precision (LRAP) scores. To make direct comparison with \cite{yin2019benchmarking} on Emotion and Situation datasets, we also report label-weighted F1 (LWF1) scores for these two tasks in Table \ref{tab:perf}. 

Here we make a special remark on the Situation and Emotion datasets: they both contain the \textit{Other} class.
For Situation this category is ``out-of-domain'' and for Emotion it is ``no emotion''. We handled the \textit{Other} classes  using the approach in Section \ref{sec:other}.

To compare all methods more fairly, we use a unified training set, testing set and class descriptions. 
Since ClassKG can achieve good results after $3$ iterations, in order to save time, we set the number of iterations to $3$.
Note that we use \textit{original} class descriptions (e.g., ``positive'', ``negative'') without rewriting them (e.g., as ``good'', ``bad'') (see Appendix \ref{app:class-name}). Therefore, the results of LOTClass, X-Class, and ClassKG are different from those reported in the original papers on tasks Amazon, Yelp, and 20News. 

In addition, because LOTClass can not find enough unlabeled documents containing class-relevant keywords as pseudo-label data on some tasks, some results are missing.
For the datasets Emotion and Situation, during the self-training process of ClassKG, extreme predictions occurred, resulting in no training data for some categories, and the training process stopped.

Although our weakly supervised method does not reach the best level, it shows good results on the whole, especially for some datasets such as SST e.g. which has a small amount of training data, our method can well solve the problem of sparse data.
In addition, our method is not highly dependent on labels, which only needs to use the original labels to achieve good results.

% These results show that variants of \textsc{Claret} are able to achieve the highest performance on each task compared with baseline methods.
% Although the best pseudo-labeling strategy depends on specific tasks, it is clear that \textsc{Claret} is overall a promising approach to dataless text classification. It performs the same as or sometimes much better than entailment models. Comparison of BM25 and \textsc{Claret} variants shows that dense retrieval module (e.g., SBERT+FAISS) is essential in obtaining pseudo-labeled documents. (See Appendix \ref{sec:appendix-performance} for supplementary performance analysis.) 

\subsection{Efficiency Comparison}%efficiency comparison
\begin{figure*}[t]
  \centering
  \subfigure[AGNews]
  {
  \begin{minipage}{.23\textwidth}
    \centering
    {\includegraphics[width=\linewidth]{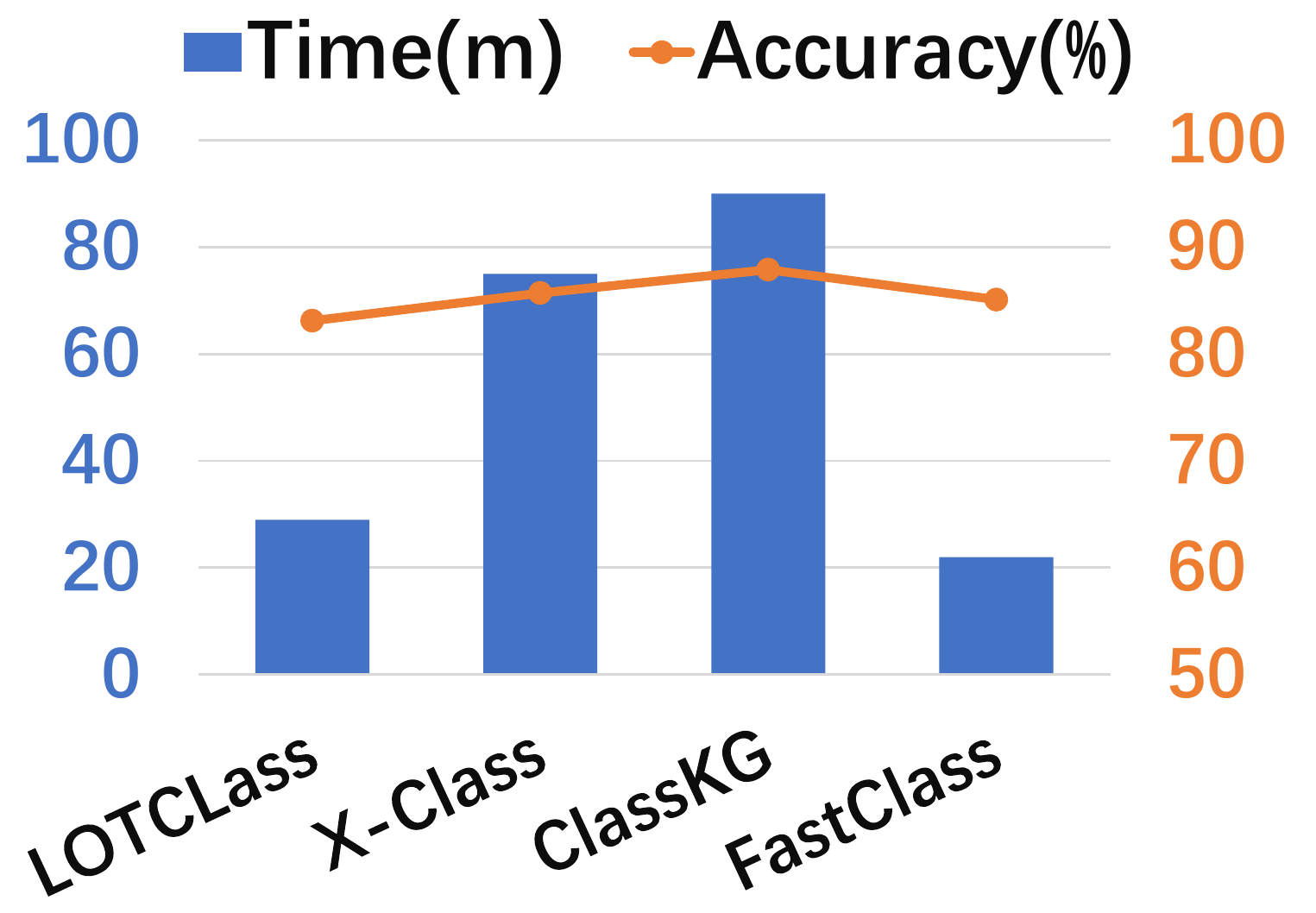}}
    \label{fig:illustrate_cases}
  \end{minipage}
  }
  \subfigure[SST]
  {
  \begin{minipage}{.23\textwidth}
    \centering
    {\includegraphics[width=\linewidth]{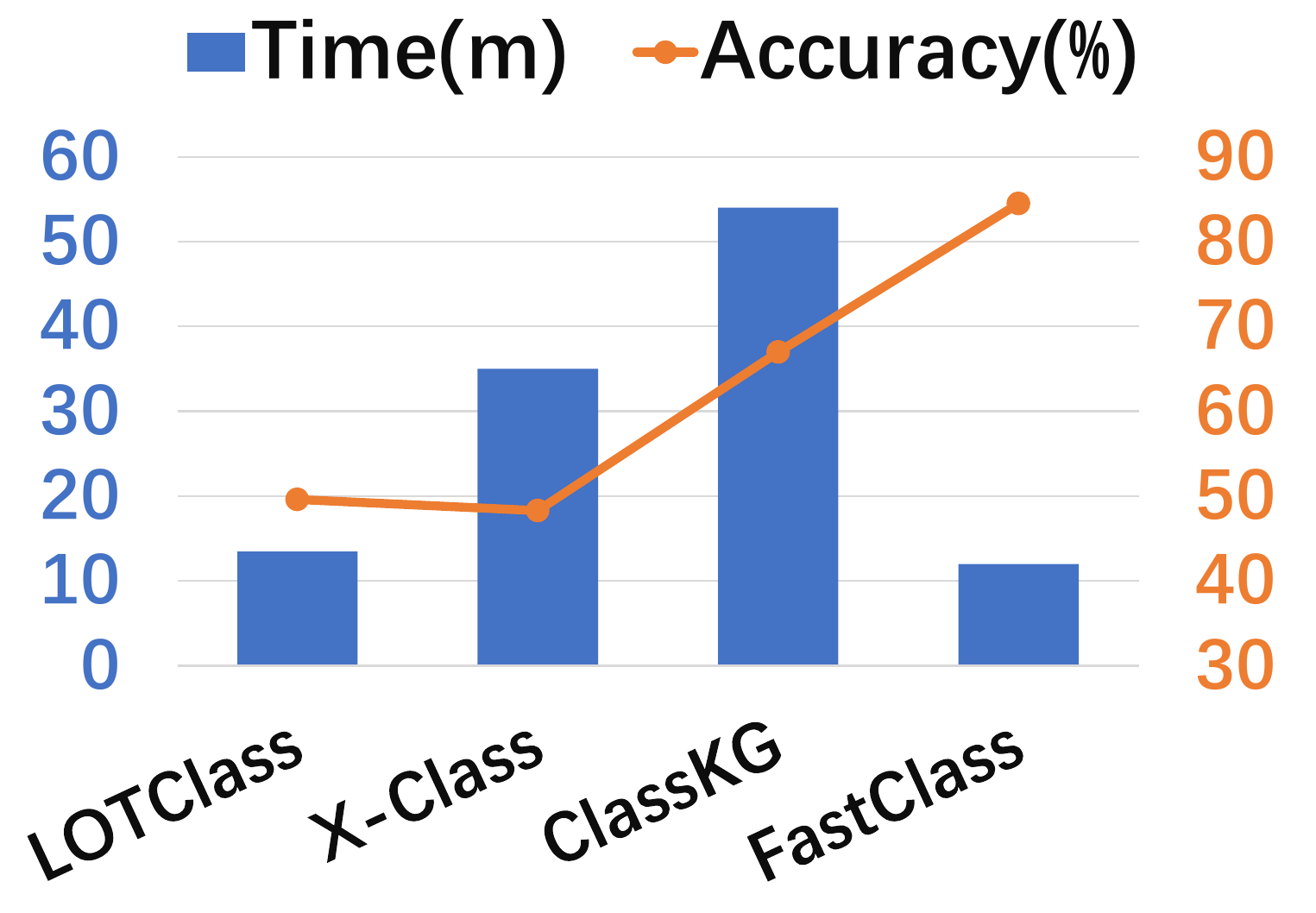}}
    \label{fig:illustrate_cases}
  \end{minipage}
  }
  \subfigure[Yahoo]
  {
  \begin{minipage}{.23\textwidth}
    \centering
    {\includegraphics[width=\linewidth]{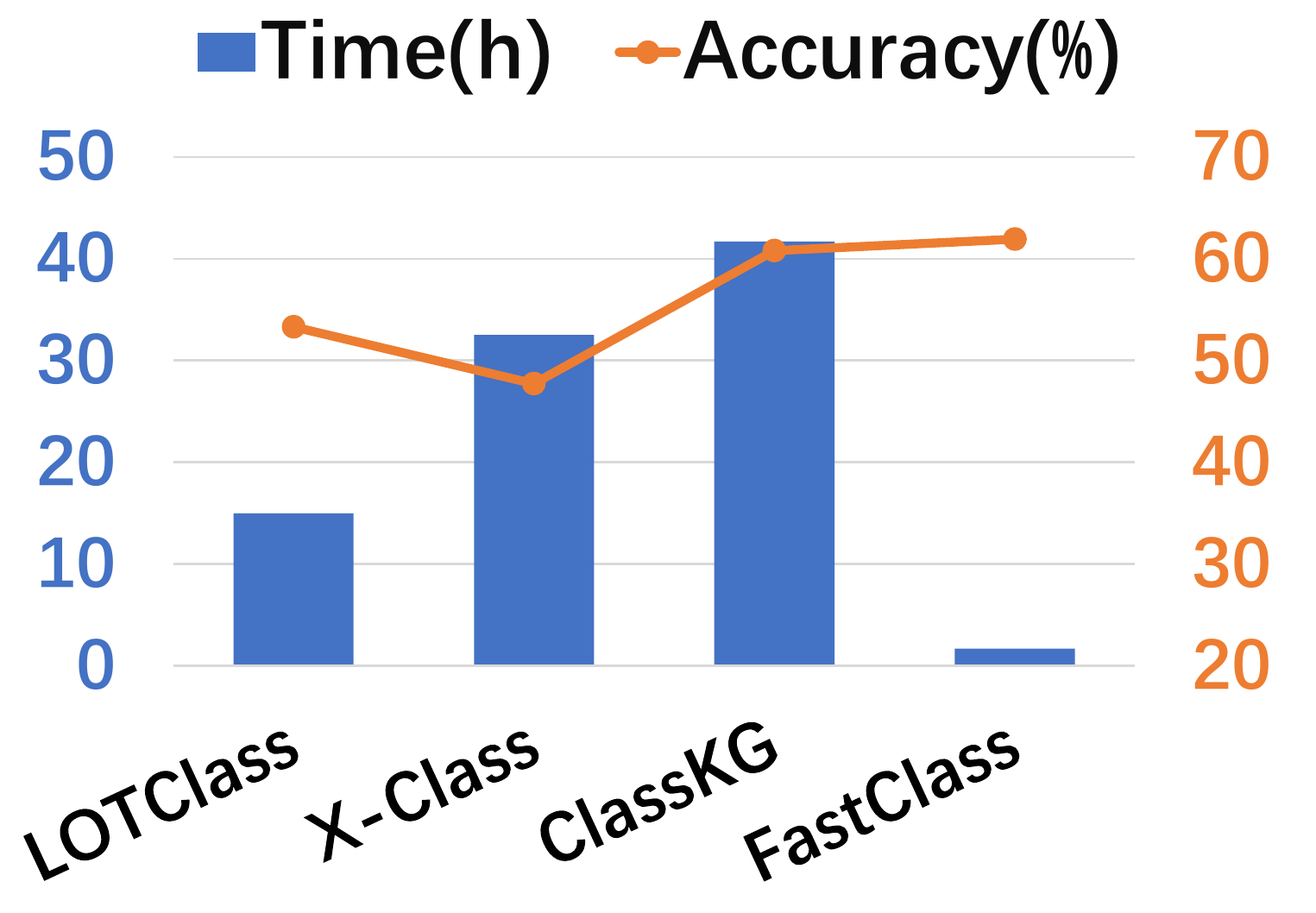}}
    \label{fig:illustrate_cases}
  \end{minipage} 
  }
  \subfigure[Amazon]
  {
  \begin{minipage}{.23\textwidth}
    \centering
    {\includegraphics[width=\linewidth]{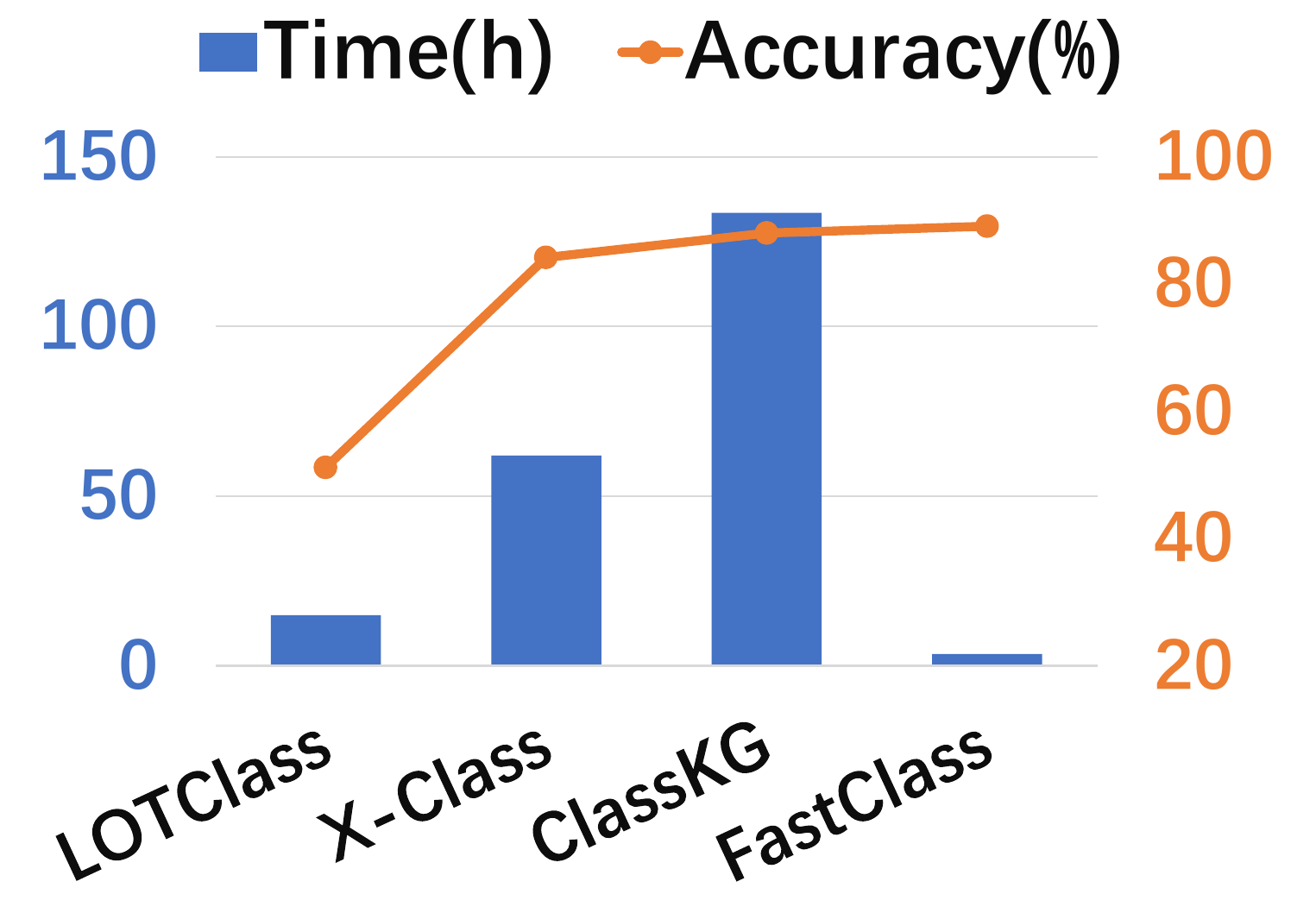}}
    \label{fig:illustrate_cases}
  \end{minipage}
  }
  \caption{Performance and efficiency comparison of weakly-supervised text classification models. Polyline: Accuracy of different models. Histogram: training time for different models.}
  \label{fig:efficiency}
  \end{figure*}

% training efficiency
\textbf{Training Efficiency}. We compare the training time of different weakly supervised methods in Table \ref{tab:efficiency}. Our method FastClass does not achieve the best results on all tasks, but FastClass can significantly save training time. Taking the dataset DBpedia as an example, the experimental results of ClassKG are significantly higher than all other methods, but its training time is more than 9 times that of FastClass. Even for the small-scale dataset SST, ClassKG takes about an hour to complete the training, but it cannot achieve the optimal results. 
For large-scale datasets, the FastClass method achieves relatively good results without sacrificing huge training time. For small-scale datasets, FastClass makes up for the problem of insufficient training due to sparse training data. Figure \ref{fig:efficiency} shows FastClass can get competitive results using the least amount of time. 

Here we need to make a point that we did not consider the computing time for sentence representation in Table \ref{tab:efficiency}, because it is a one-time, once-and-for-all process. The sentence embedding process for the entire external document repository took less than an hour, and we provided the Python pickle file of the embedding as part of the data and code release for other researchers to use in our Github repository.
\begin{table}[h]
\renewcommand\arraystretch{1} 
\renewcommand\tabcolsep{5pt}
	\centering
	\begin{tabular}{ccc}
    \toprule
	\textbf{Method}	&\textbf{Total Time} &\textbf{Per Document}\\
	\midrule
	
    0SHOT-TC &2162.4s &22ms\\

    NATCAT &1485.8s &15ms\\

    FastClass &\textbf{\underline{306.7s}} &\textbf{\underline{3ms}}\\
    \toprule
	\end{tabular}
\caption{Total testing time on Yahoo using \textit{label-fully-unseen} 0SHOT-TC, NATCAT and FastClass. All methods used RoBERTa-base model.}
\label{tab:pred-time}
\end{table}

\textbf{Prediction efficiency}.
 Compare with the commonly used entailment model which only need to be trained once to be applied to any task in dataless tasks. A big advantage of classification models over entailment models is the prediction speed. Classification models only need one forward pass to make a prediction for $k$ categories, whereas entailment models need $k$ forward passes. Table \ref{tab:pred-time} compares prediction time of entailment models and FastClass on the Yahoo dataset (100,000 documents). Our method is not only more accurate (Table \ref{tab:perf}) but also 5-7 times faster.

\subsection{Ablation Studies}
\textbf{Seed sets}: To study the benefit of using task-specific unlabeled data as ``seed sets'', we directly construct a pseudo training set based on the similarity between the class description and the external documents. We call this variant FastClass$_{{external}}$. It directly uses the class descriptions as queries to retrieve data from external data. 
For class $j$, we retrieve $n'$ most relevant documents from  external data with respect to the class description $d_j$, $R_j = \{x_i\}_{i=1}^{n'}$.
Then, we use the retrieved external data for subset selection, which is consistent with FastClass include subset diversification and entropy maximization. The parameter $\theta$ is $n' = \{2m, 5m, 10m\}$ and $m = \{300, 500, 800\}$.
\begin{table}[h]
  \renewcommand\arraystretch{1}
    \renewcommand\tabcolsep{2.5pt}
	\centering
	\begin{tabular}{l|cccc}
	\toprule
	Method & AGnews &Dbpedia &SST &Yelp\\
	\midrule
    FastClass$_{external}$ &72.7 &84.9 &80.1 &\textbf{87.9}\\
    
    FastClass &\textbf{85.1} &\textbf{92.4} &\textbf{84.5} &\textbf{87.9} \\

    \bottomrule

	\end{tabular}
	\caption{Impact of seed sets on performance (\%). The best performance of each column is in bold.} 
	\label{external}
\end{table}

Table \ref{external} shows that instead of using task specific data as retrieval queries, directly using class descriptions for retrieval will affect the experimental results. Although there is a semantic gap between task data and class descriptions, task data can represent the textual features of specific task, such as text style, which is beneficial for retrieving semantically related and stylistically similar documents.

\textbf{Facility location function}: To study the impact of the facility location subset selection function on the diversity of pseudo-labeled 
documents, we conduct an ablation experiment. We remove the facility location function and  directly select $m$ documents based on the similarity score between external documents and seed documents. The other steps are consistent with FastClass, i.e., it also uses maximum entropy to select the optimal subset. The comparison  results are shown in Table \ref{facility}. We selected four representative tasks: AGnews, DBpedia, Yelp, and SST. The results confirmed our hypothesis: facility location subset selection helped improve model performance in most cases by choosing a more diverse data subset.

\begin{table}[h]

  \renewcommand\arraystretch{1}
    \renewcommand\tabcolsep{2.5pt}
	\centering
	\begin{tabular}{p{2cm}|m{1.2cm}<{\centering}m{1.2cm}<{\centering}m{1.2cm}<{\centering}m{1.2cm}}
	
	\toprule
	Method & AGnews &Dbpedia &SST &Yelp\\
	\midrule
    FastClass$_{wo}$ &84.7 &90.3 &80.4 &83.1\\
    
    FastClass &\textbf{85.1} &\textbf{92.4} &\textbf{84.5} &\textbf{87.9} \\

    \bottomrule

	\end{tabular}
	\caption{Impact of the facility location subset selection on  model performance (\%). The best performance of each column is in bold. FastClass$_{wo}$: FastClass without facility location subset selection.} 
	\label{facility}
\end{table}

\vspace{-.2in}
\section{Conclusion}
We proposed a weakly-supervised text classification method FastClass which selects a document subset returned by dense retrieval models for classifier training.
Compared to keyword-driven methods, since our approach no longer needs to expand each class description into a set of class-specific keywords,  it is less reliant on initial class descriptions.
Extensive experiments show that the proposed method is able to achieve competitive classification performance which does not require high-quality initial class descriptions and often saves significant training time.

% This work opens up many interesting paths for future work. First, although our methods achieved good performance relative to baseline methods, the absolute performance is still low on some datasets, e.g., Emotion, Situation, and Comment. Future work should focus on these challenging scenarios where many target categories exist, each document can have multiple labels, and task-specific data are scarce. Second, our methods implies that it is possible to warm-start a classifier using class descriptions only. This means in an interactive machine learning setting, a user can have an up-and-running classifier  warm-started classifier can save the initial effort in labeling an initial set of labels. 

\section*{Acknowledgement}
The authors would like to thank the anonymous referees for their valuable comments. This work is supported by the  National Natural Science Foundation of China (No. 61976102 and No. U19A2065), the Science and Technology Development Program of Jilin Province (No.20210508060RQ), and the Fundamental Research Funds for the Central Universities, JLU.

\section*{Limitations}
The limitations of our method are mainly the following.
First, we propose a practically efficient approach without using sophisticated architecture, so the technical approach is relatively simple.
Second, the performance of FastClass is competitive on all tasks but only achieves the best results on some tasks.

\bibliography{anthology,custom}

\begin{thebibliography}{44}
\expandafter\ifx\csname natexlab\endcsname\relax\def\natexlab#1{#1}\fi

\bibitem[{Arase et~al.(2021)Arase, Blunsom, Diab, Dodge, Gurevych, Liang,
  Raffel, R\"{u}ckl\'{e}, Schwartz, Smith, Strubell, and
  Zhang}]{arase2021efficient}
Yuki Arase, Phil Blunsom, Mona Diab, Jesse Dodge, Iryna Gurevych, Percy Liang,
  Colin Raffel, Andreas R\"{u}ckl\'{e}, Roy Schwartz, Noah~A. Smith, Emma
  Strubell, and Yue Zhang. 2021.
\newblock \href
  {https://www.aclweb.org/portal/content/efficient-nlp-policy-document}
  {Efficient nlp policy document}.

\bibitem[{Baram et~al.(2004)Baram, Yaniv, and Luz}]{baram2004online}
Yoram Baram, Ran~El Yaniv, and Kobi Luz. 2004.
\newblock Online choice of active learning algorithms.
\newblock \emph{Journal of Machine Learning Research}, 5(Mar):255--291.

\bibitem[{Bhatia et~al.(2016)Bhatia, Dahiya, Jain, Kar, Mittal, Prabhu, and
  Varma}]{Bhatia16}
K.~Bhatia, K.~Dahiya, H.~Jain, P.~Kar, A.~Mittal, Y.~Prabhu, and M.~Varma.
  2016.
\newblock \href {http://manikvarma.org/downloads/XC/XMLRepository.html} {The
  extreme classification repository: Multi-label datasets and code}.

\bibitem[{Blitzer et~al.(2007)Blitzer, Dredze, Pereira, and
  Biographies}]{blitzer2007boom}
John Blitzer, Mark Dredze, Fernando Pereira, and Bollywood Biographies. 2007.
\newblock Boom-boxes and blenders: domain adaptation for sentiment
  classification.
\newblock In \emph{Proceedings of the 45th Annual Meeting of the Association
  for Computational Linguistics (ACL’07), Pereira}, volume 447.

\bibitem[{Chang et~al.(2008)Chang, Ratinov, Roth, and
  Srikumar}]{chang2008importance}
Ming-Wei Chang, Lev-Arie Ratinov, Dan Roth, and Vivek Srikumar. 2008.
\newblock Importance of semantic representation: Dataless classification.
\newblock In \emph{Aaai}, volume~2, pages 830--835.

\bibitem[{Chu et~al.(2020)Chu, Stratos, and Gimpel}]{chu2020natcat}
Zewei Chu, Karl Stratos, and Kevin Gimpel. 2020.
\newblock Natcat: Weakly supervised text classification with naturally
  annotated datasets.
\newblock \emph{arXiv preprint arXiv:2009.14335}.

\bibitem[{Devlin et~al.(2018)Devlin, Chang, Lee, and
  Toutanova}]{devlin2018bert}
Jacob Devlin, Ming-Wei Chang, Kenton Lee, and Kristina Toutanova. 2018.
\newblock Bert: Pre-training of deep bidirectional transformers for language
  understanding.
\newblock \emph{arXiv preprint arXiv:1810.04805}.

\bibitem[{Jaynes(1957)}]{jaynes1957information}
Edwin~T Jaynes. 1957.
\newblock Information theory and statistical mechanics.
\newblock \emph{Physical review}, 106(4):620.

\bibitem[{Johnson et~al.(2017)Johnson, Douze, and J{\'e}gou}]{JDH17}
Jeff Johnson, Matthijs Douze, and Herv{\'e} J{\'e}gou. 2017.
\newblock Billion-scale similarity search with gpus.
\newblock \emph{arXiv preprint arXiv:1702.08734}.

\bibitem[{Lang(1995)}]{lang1995newsweeder}
Ken Lang. 1995.
\newblock Newsweeder: Learning to filter netnews.
\newblock In \emph{Machine Learning Proceedings 1995}, pages 331--339.
  Elsevier.

\bibitem[{Lehmann et~al.(2015)Lehmann, Isele, Jakob, Jentzsch, Kontokostas,
  Mendes, Hellmann, Morsey, Van~Kleef, Auer et~al.}]{lehmann2015dbpedia}
Jens Lehmann, Robert Isele, Max Jakob, Anja Jentzsch, Dimitris Kontokostas,
  Pablo~N Mendes, Sebastian Hellmann, Mohamed Morsey, Patrick Van~Kleef,
  S{\"o}ren Auer, et~al. 2015.
\newblock Dbpedia--a large-scale, multilingual knowledge base extracted from
  wikipedia.
\newblock \emph{Semantic web}, 6(2):167--195.

\bibitem[{Li et~al.(2018)Li, Li, Chi, Ouyang, and Li}]{li2018dataless}
Ximing Li, Changchun Li, Jinjin Chi, Jihong Ouyang, and Chenliang Li. 2018.
\newblock Dataless text classification: A topic modeling approach with document
  manifold.
\newblock In \emph{Proceedings of the 27th ACM International Conference on
  Information and Knowledge Management}, pages 973--982.

\bibitem[{Li and Yang(2018)}]{li2018pseudo}
Ximing Li and Bo~Yang. 2018.
\newblock A pseudo label based dataless naive bayes algorithm for text
  classification with seed words.
\newblock In \emph{Proceedings of the 27th International Conference on
  Computational Linguistics}, pages 1908--1917.

\bibitem[{Liu et~al.(2019{\natexlab{a}})Liu, Zhang, Fan, Fu, Li, Wu, and
  Lam}]{liu2019reconstructing}
Han Liu, Xiaotong Zhang, Lu~Fan, Xuandi Fu, Qimai Li, Xiao-Ming Wu, and
  Albert~YS Lam. 2019{\natexlab{a}}.
\newblock Reconstructing capsule networks for zero-shot intent classification.
\newblock In \emph{Proceedings of the 2019 Conference on Empirical Methods in
  Natural Language Processing and the 9th International Joint Conference on
  Natural Language Processing (EMNLP-IJCNLP)}, pages 4799--4809.

\bibitem[{Liu et~al.(2019{\natexlab{b}})Liu, Ott, Goyal, Du, Joshi, Chen, Levy,
  Lewis, Zettlemoyer, and Stoyanov}]{liu2019roberta}
Yinhan Liu, Myle Ott, Naman Goyal, Jingfei Du, Mandar Joshi, Danqi Chen, Omer
  Levy, Mike Lewis, Luke Zettlemoyer, and Veselin Stoyanov. 2019{\natexlab{b}}.
\newblock Roberta: A robustly optimized bert pretraining approach.
\newblock \emph{arXiv preprint arXiv:1907.11692}.

\bibitem[{Lo et~al.(2020)Lo, Wang, Neumann, Kinney, and
  Weld}]{lo-wang-2020-s2orc}
Kyle Lo, Lucy~Lu Wang, Mark Neumann, Rodney Kinney, and Daniel Weld. 2020.
\newblock \href {https://doi.org/10.18653/v1/2020.acl-main.447} {{S}2{ORC}: The
  semantic scholar open research corpus}.
\newblock In \emph{Proceedings of the 58th Annual Meeting of the Association
  for Computational Linguistics}, pages 4969--4983, Online. Association for
  Computational Linguistics.

\bibitem[{Loshchilov and Hutter(2017)}]{loshchilov2017decoupled}
Ilya Loshchilov and Frank Hutter. 2017.
\newblock Decoupled weight decay regularization.
\newblock \emph{arXiv preprint arXiv:1711.05101}.

\bibitem[{Mekala and Shang(2020)}]{mekala2020contextualized}
Dheeraj Mekala and Jingbo Shang. 2020.
\newblock Contextualized weak supervision for text classification.
\newblock In \emph{Proceedings of the 58th Annual Meeting of the Association
  for Computational Linguistics}, pages 323--333.

\bibitem[{Meng et~al.(2018)Meng, Shen, Zhang, and Han}]{meng2018weakly}
Yu~Meng, Jiaming Shen, Chao Zhang, and Jiawei Han. 2018.
\newblock Weakly-supervised neural text classification.
\newblock In \emph{proceedings of the 27th ACM International Conference on
  information and knowledge management}, pages 983--992.

\bibitem[{Meng et~al.(2019)Meng, Shen, Zhang, and Han}]{meng2019weakly}
Yu~Meng, Jiaming Shen, Chao Zhang, and Jiawei Han. 2019.
\newblock Weakly-supervised hierarchical text classification.
\newblock In \emph{Proceedings of the AAAI conference on artificial
  intelligence}, pages 6826--6833.

\bibitem[{Meng et~al.(2020)Meng, Zhang, Huang, Xiong, Ji, Zhang, and
  Han}]{meng2020text}
Yu~Meng, Yunyi Zhang, Jiaxin Huang, Chenyan Xiong, Heng Ji, Chao Zhang, and
  Jiawei Han. 2020.
\newblock Text classification using label names only: A language model
  self-training approach.
\newblock In \emph{EMNLP (1)}.

\bibitem[{Nam et~al.(2016)Nam, Menc{\'\i}a, and F{\"u}rnkranz}]{nam2016all}
Jinseok Nam, Eneldo~Loza Menc{\'\i}a, and Johannes F{\"u}rnkranz. 2016.
\newblock All-in text: Learning document, label, and word representations
  jointly.
\newblock In \emph{Thirtieth AAAI Conference on Artificial Intelligence}.

\bibitem[{Nemhauser et~al.(1978)Nemhauser, Wolsey, and
  Fisher}]{nemhauser1978analysis}
George~L Nemhauser, Laurence~A Wolsey, and Marshall~L Fisher. 1978.
\newblock An analysis of approximations for maximizing submodular set
  functions—i.
\newblock \emph{Mathematical programming}, 14(1):265--294.

\bibitem[{Oberl{\"a}nder and Klinger(2018)}]{oberlander2018analysis}
Laura Ana~Maria Oberl{\"a}nder and Roman Klinger. 2018.
\newblock An analysis of annotated corpora for emotion classification in text.
\newblock In \emph{Proceedings of the 27th International Conference on
  Computational Linguistics}, pages 2104--2119.

\bibitem[{Pushp and Srivastava(2017)}]{pushp2017train}
Pushpankar~Kumar Pushp and Muktabh~Mayank Srivastava. 2017.
\newblock Train once, test anywhere: Zero-shot learning for text
  classification.
\newblock \emph{arXiv preprint arXiv:1712.05972}.

\bibitem[{Reimers and Gurevych(2019)}]{reimers-2019-sentence-bert}
Nils Reimers and Iryna Gurevych. 2019.
\newblock \href {https://arxiv.org/abs/1908.10084} {Sentence-bert: Sentence
  embeddings using siamese bert-networks}.
\newblock In \emph{Proceedings of the 2019 Conference on Empirical Methods in
  Natural Language Processing}. Association for Computational Linguistics.

\bibitem[{Rios and Kavuluru(2018)}]{rios2018few}
Anthony Rios and Ramakanth Kavuluru. 2018.
\newblock Few-shot and zero-shot multi-label learning for structured label
  spaces.
\newblock In \emph{Proceedings of the Conference on Empirical Methods in
  Natural Language Processing. Conference on Empirical Methods in Natural
  Language Processing}, volume 2018, page 3132. NIH Public Access.

\bibitem[{Schreiber et~al.(2020)Schreiber, Bilmes, and
  Noble}]{schreiber2020apricot}
Jacob~M Schreiber, Jeffrey~A Bilmes, and William~Stafford Noble. 2020.
\newblock apricot: Submodular selection for data summarization in python.
\newblock \emph{J. Mach. Learn. Res.}, 21:161--1.

\bibitem[{Schwartz et~al.(2020)Schwartz, Dodge, Smith, and
  Etzioni}]{schwartz2020green}
Roy Schwartz, Jesse Dodge, Noah~A Smith, and Oren Etzioni. 2020.
\newblock Green ai.
\newblock \emph{Communications of the ACM}, 63(12):54--63.

\bibitem[{Shen et~al.(2021)Shen, Qiu, Meng, Shang, Ren, and
  Han}]{shen2021taxoclass}
Jiaming Shen, Wenda Qiu, Yu~Meng, Jingbo Shang, Xiang Ren, and Jiawei Han.
  2021.
\newblock Taxoclass: Hierarchical multi-label text classification using only
  class names.
\newblock In \emph{Proceedings of the 2021 Conference of the North American
  Chapter of the Association for Computational Linguistics: Human Language
  Technologies}, pages 4239--4249.

\bibitem[{Socher et~al.(2013)Socher, Perelygin, Wu, Chuang, Manning, Ng, and
  Potts}]{socher2013recursive}
Richard Socher, Alex Perelygin, Jean Wu, Jason Chuang, Christopher~D Manning,
  Andrew~Y Ng, and Christopher Potts. 2013.
\newblock Recursive deep models for semantic compositionality over a sentiment
  treebank.
\newblock In \emph{Proceedings of the 2013 conference on empirical methods in
  natural language processing}, pages 1631--1642.

\bibitem[{Song and Roth(2014)}]{song2014dataless}
Yangqiu Song and Dan Roth. 2014.
\newblock On dataless hierarchical text classification.
\newblock In \emph{Twenty-eighth AAAI conference on artificial intelligence}.

\bibitem[{Thakur et~al.(2021)Thakur, Reimers, R{\"u}ckl{\'e}, Srivastava, and
  Gurevych}]{thakur2021beir}
Nandan Thakur, Nils Reimers, Andreas R{\"u}ckl{\'e}, Abhishek Srivastava, and
  Iryna Gurevych. 2021.
\newblock Beir: A heterogenous benchmark for zero-shot evaluation of
  information retrieval models.
\newblock \emph{arXiv preprint arXiv:2104.08663}.

\bibitem[{Wang et~al.(2019)Wang, Zheng, Yu, and Miao}]{wang2019survey}
Wei Wang, Vincent~W Zheng, Han Yu, and Chunyan Miao. 2019.
\newblock A survey of zero-shot learning: Settings, methods, and applications.
\newblock \emph{ACM Transactions on Intelligent Systems and Technology (TIST)},
  10(2):1--37.

\bibitem[{Wang et~al.(2021)Wang, Mekala, and Shang}]{wang2021x}
Zihan Wang, Dheeraj Mekala, and Jingbo Shang. 2021.
\newblock X-class: Text classification with extremely weak supervision.
\newblock In \emph{Proceedings of the 2021 Conference of the North American
  Chapter of the Association for Computational Linguistics: Human Language
  Technologies}, pages 3043--3053.

\bibitem[{Wei et~al.(2015)Wei, Iyer, and Bilmes}]{wei2015submodularity}
Kai Wei, Rishabh Iyer, and Jeff Bilmes. 2015.
\newblock Submodularity in data subset selection and active learning.
\newblock In \emph{International Conference on Machine Learning}, pages
  1954--1963. PMLR.

\bibitem[{Wu et~al.(2020)Wu, Qiao, Chen, Wu, Qi, Lian, Liu, Xie, Gao, Wu
  et~al.}]{wu2020mind}
Fangzhao Wu, Ying Qiao, Jiun-Hung Chen, Chuhan Wu, Tao Qi, Jianxun Lian,
  Danyang Liu, Xing Xie, Jianfeng Gao, Winnie Wu, et~al. 2020.
\newblock Mind: A large-scale dataset for news recommendation.
\newblock In \emph{Proceedings of the 58th Annual Meeting of the Association
  for Computational Linguistics}, pages 3597--3606.

\bibitem[{Xia et~al.(2018)Xia, Zhang, Yan, Chang, and Yu}]{xia2018zero}
Congying Xia, Chenwei Zhang, Xiaohui Yan, Yi~Chang, and Philip~S Yu. 2018.
\newblock Zero-shot user intent detection via capsule neural networks.
\newblock \emph{arXiv preprint arXiv:1809.00385}.

\bibitem[{Ye et~al.(2020)Ye, Geng, Chen, Chen, Xu, Zheng, Wang, Zhang, and
  Chen}]{ye2020zero}
Zhiquan Ye, Yuxia Geng, Jiaoyan Chen, Jingmin Chen, Xiaoxiao Xu, SuHang Zheng,
  Feng Wang, Jun Zhang, and Huajun Chen. 2020.
\newblock Zero-shot text classification via reinforced self-training.
\newblock In \emph{Proceedings of the 58th Annual Meeting of the Association
  for Computational Linguistics}, pages 3014--3024.

\bibitem[{Yin et~al.(2019)Yin, Hay, and Roth}]{yin2019benchmarking}
Wenpeng Yin, Jamaal Hay, and Dan Roth. 2019.
\newblock Benchmarking zero-shot text classification: Datasets, evaluation and
  entailment approach.
\newblock In \emph{Proceedings of the 2019 Conference on Empirical Methods in
  Natural Language Processing and the 9th International Joint Conference on
  Natural Language Processing (EMNLP-IJCNLP)}, pages 3914--3923.

\bibitem[{Zellers et~al.(2019)Zellers, Holtzman, Rashkin, Bisk, Farhadi,
  Roesner, and Choi}]{zellers2019grover}
Rowan Zellers, Ari Holtzman, Hannah Rashkin, Yonatan Bisk, Ali Farhadi,
  Franziska Roesner, and Yejin Choi. 2019.
\newblock Defending against neural fake news.
\newblock In \emph{Advances in Neural Information Processing Systems 32}.

\bibitem[{Zhang et~al.(2019)Zhang, Lertvittayakumjorn, and
  Guo}]{zhang2019integrating}
Jingqing Zhang, Piyawat Lertvittayakumjorn, and Yike Guo. 2019.
\newblock Integrating semantic knowledge to tackle zero-shot text
  classification.
\newblock In \emph{Proceedings of NAACL-HLT}, pages 1031--1040.

\bibitem[{Zhang et~al.(2021)Zhang, Ding, Xu, Liu, and Zhou}]{zhang2021weakly}
Lu~Zhang, Jiandong Ding, Yi~Xu, Yingyao Liu, and Shuigeng Zhou. 2021.
\newblock Weakly-supervised text classification based on keyword graph.
\newblock In \emph{Proceedings of the 2021 Conference on Empirical Methods in
  Natural Language Processing}, pages 2803--2813.

\bibitem[{Zhang et~al.(2015)Zhang, Zhao, and LeCun}]{zhang2015character}
Xiang Zhang, Junbo Zhao, and Yann LeCun. 2015.
\newblock Character-level convolutional networks for text classification.
\newblock \emph{Advances in neural information processing systems},
  28:649--657.

\end{thebibliography}
\bibliographystyle{acl_natbib}

\newpage
\appendix

\section{Class Descriptions in Evaluation Datasets}
\label{app:class-name}
We list the class descriptions of the datasets we used for evaluation as follows. These texts are used as to compute SBERT  vector representations. %Note that some class descriptions are very abstract: ``positive'' and ``negative'' for sentiment classification datasets (Yelp, Amazon, SST-B).

\textbf{Yahoo}: Society and Culture; Science and Mathematics; Health, Education and Reference; Computers and Internet; Sports; Business and Finance; Entertainment and Music; Family and Relationships; Politics and Government.

\textbf{AGnews}: politics; sports; business; technology.

\textbf{20Newsgroup}: atheist atheism;
        computer graphics;
        computer OS microsoft windows miscellaneous;
        computer system IBM PC hardware;
        computer system Mac hardware;
        computer windows;
        miscellaneous for sale;
        recreational automobile;
        recreational motorcycles;
        recreational sport baseball;
        recreational sport hockey;
        science  cryptography;
        science electronics;
        science medicine;
        science space;
        society religion christian;
        talk politics guns;
        talk politics middle East;  
        talk politics miscellaneous;
        talk religion miscellaneous.

\textbf{DBPedia}: Company;
                Educational Institution;
                Artist;
                Athlete;
                Office Holder;
                Mean Of Transportation;
                Building;
                Natural Place;
                Village;
                Animal;
                Plant;
                Album;
                Film;
                Written Work.
                
\textbf{Yelp}: positive; negative.

\textbf{Amazon}: positive; negative.

\textbf{SST-B}: positive; negative.

\textbf{Emotion}: anger;
                sadness;
                surprise;
                love;
                fear;
                disgust;
                guilt;
                shame;
                joy;
                no emotion.

\textbf{Situation}: utilities energy or sanitation;
                water supply;
                search/rescue;
                medical assistance;
                infrastructure;
                shelter;
                evacuation;
                regime change;
                food supply;
                crime violence;
                terrorism;
                out-of-domain.

\section{BERT-based Classifier Performance}
We have reported the results based on RoBERTa as our main result. Here we show the classification performance based on  BERT classifier in Table~\ref{tab:bert_result}. In most cases, we found that the performance of RoBERTa model is better than  BERT. This may be because compared with BERT's use of Wikipedia and books the training data of RoBERTa comes from web text which is more diverse. 

\begin{table*}[htbp]
\renewcommand\arraystretch{1} 
\renewcommand\tabcolsep{4pt}
	\centering
	\begin{tabular}{l|ccccccccc}
	\toprule
	Method &Yahoo &AGnews &20News &DBPedia &Yelp &Emotion &Amazon &SST  &Situation \\
	\midrule
    FastClass$_{Bert}$ &60.8	&83.9	&53.9	&92.8	&86.7  &26.8 &84.1	&81.8	&38.0\\ 

    FastClass$_{RoBerta}$ &61.9	&85.1	&55.4	&92.4	&87.9  &28.0 &89.1	&84.5	&39.2\\ 
	\bottomrule
	\end{tabular}
	\caption{Weakly-supervised text classification performance in nine datasets (\%) based on BERT and RoBERTa classifier.} 
\label{tab:bert_result}
\end{table*}

\section{Relation Between Entropy and Accuracy}
\label{sec:ent-acc-relation}
\begin{figure*}[bp]
 \centering
 \subfigure
  {
   \begin{minipage}{.32\textwidth}
    \centerline
    {\includegraphics[width=\linewidth]{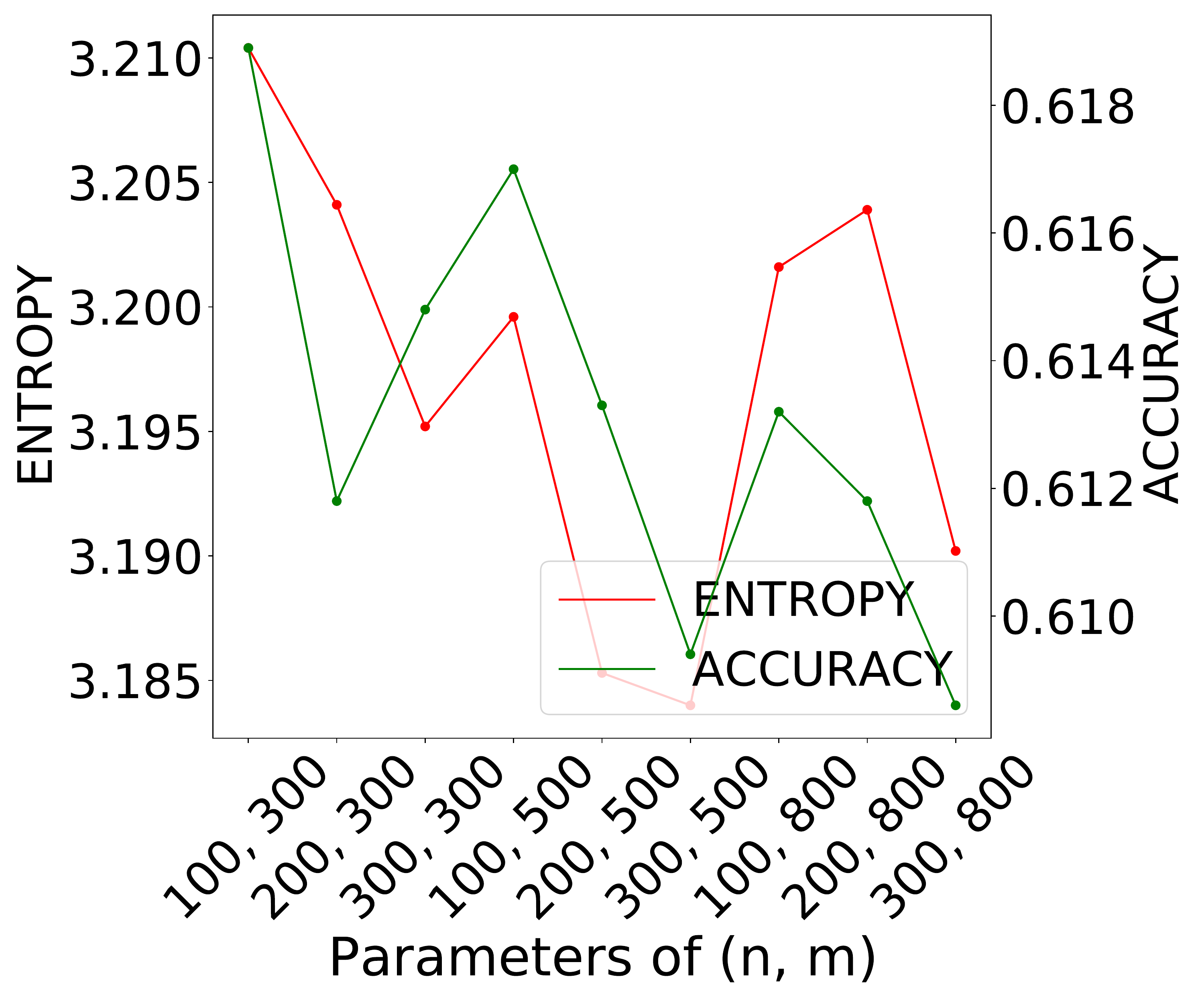}}
    \centerline{\small Yahoo}
   \end{minipage}
   \begin{minipage}{.32\textwidth}
    \centerline
    {\includegraphics[width=\linewidth]{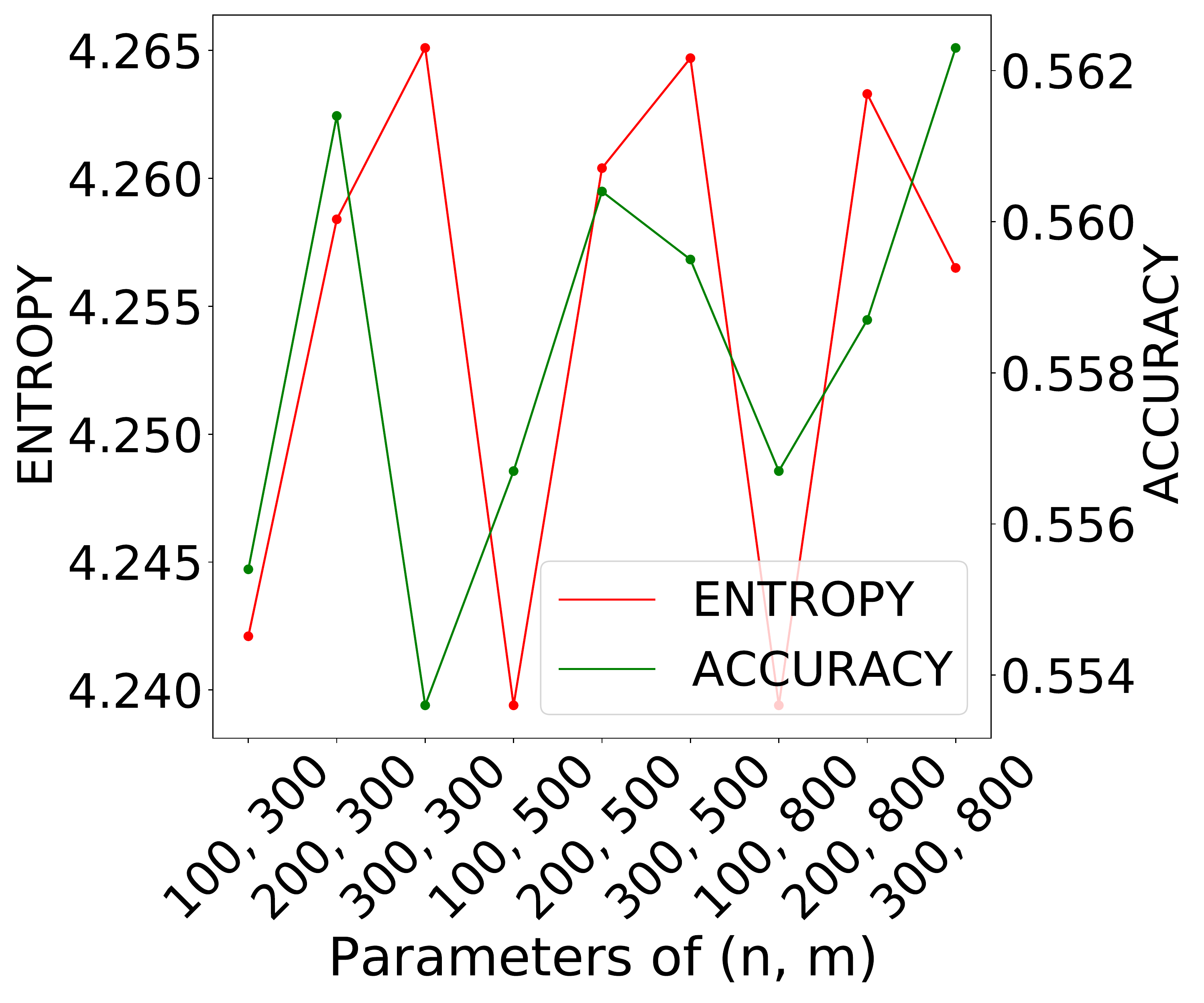}}
    \centerline{\small 20News}
   \end{minipage}
   \begin{minipage}{.32\textwidth}
    \centerline
    {\includegraphics[width=\linewidth]{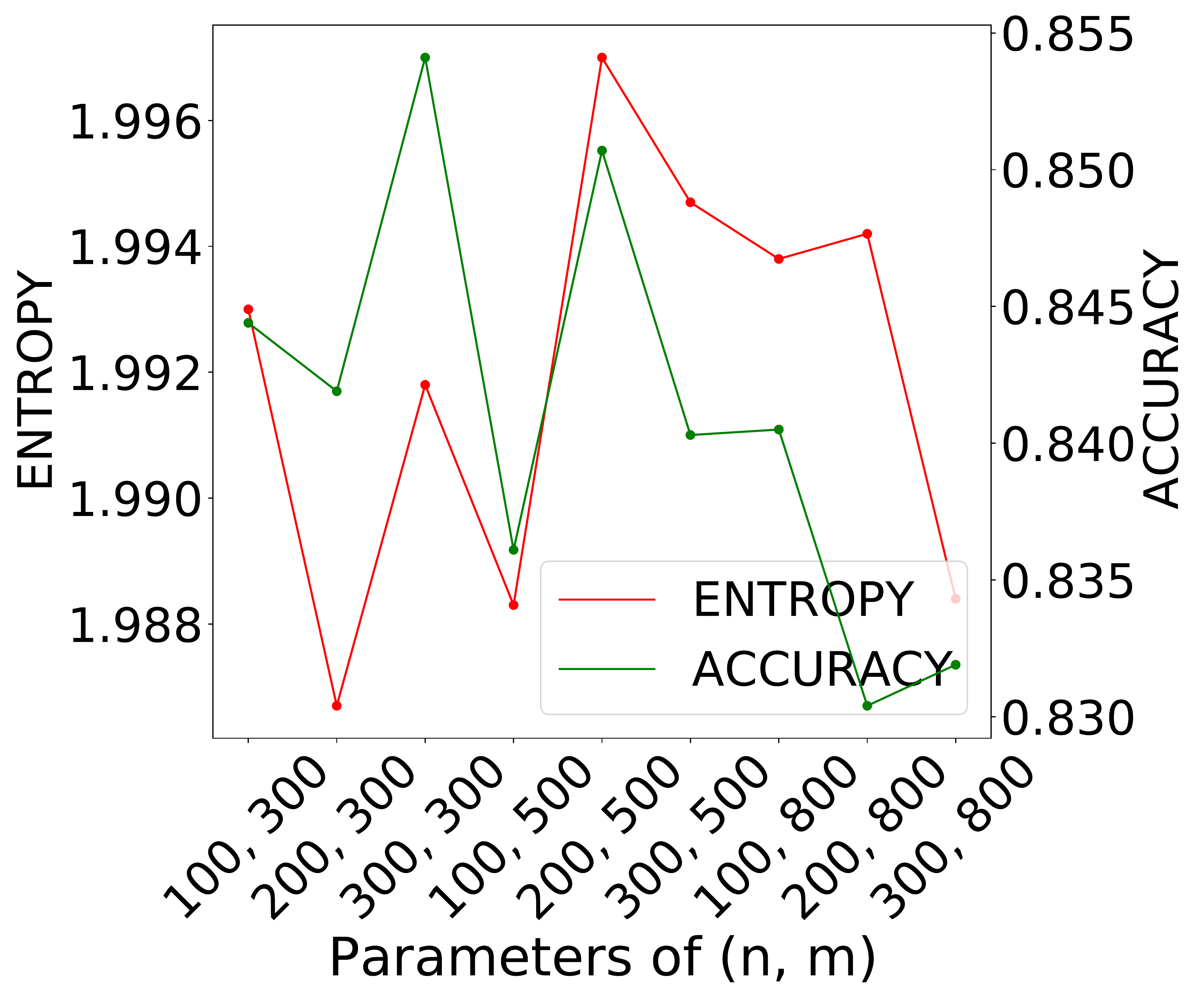}}
    \centerline{\small AGNews}
   \end{minipage}
  }

  \subfigure
  {
   \begin{minipage}{.32\textwidth}
    \centerline
    {\includegraphics[width=\linewidth]{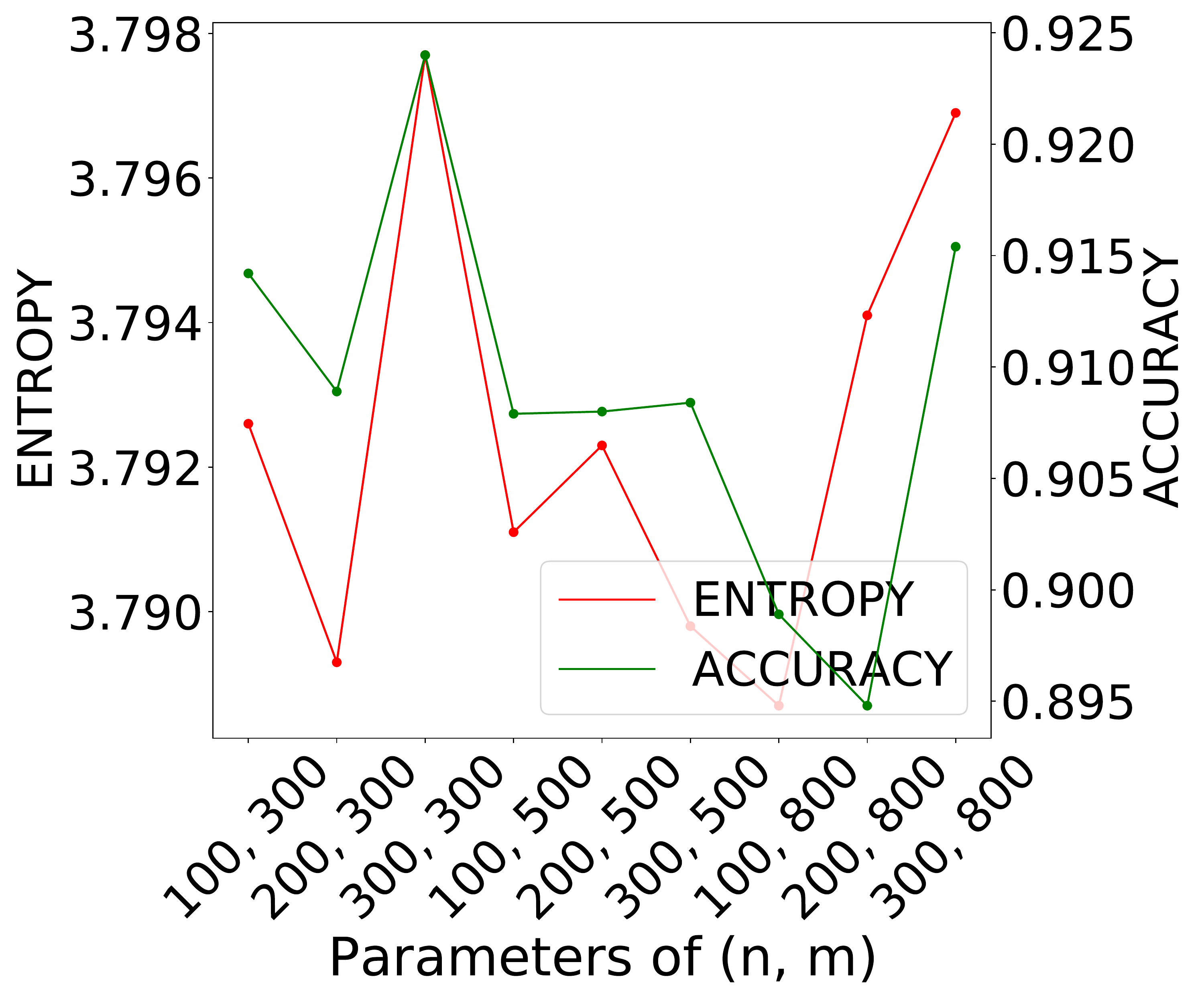}}
    \centerline{\small DBPedia}
   \end{minipage}
   \begin{minipage}{.32\textwidth}
    \centerline
    {\includegraphics[width=\linewidth]{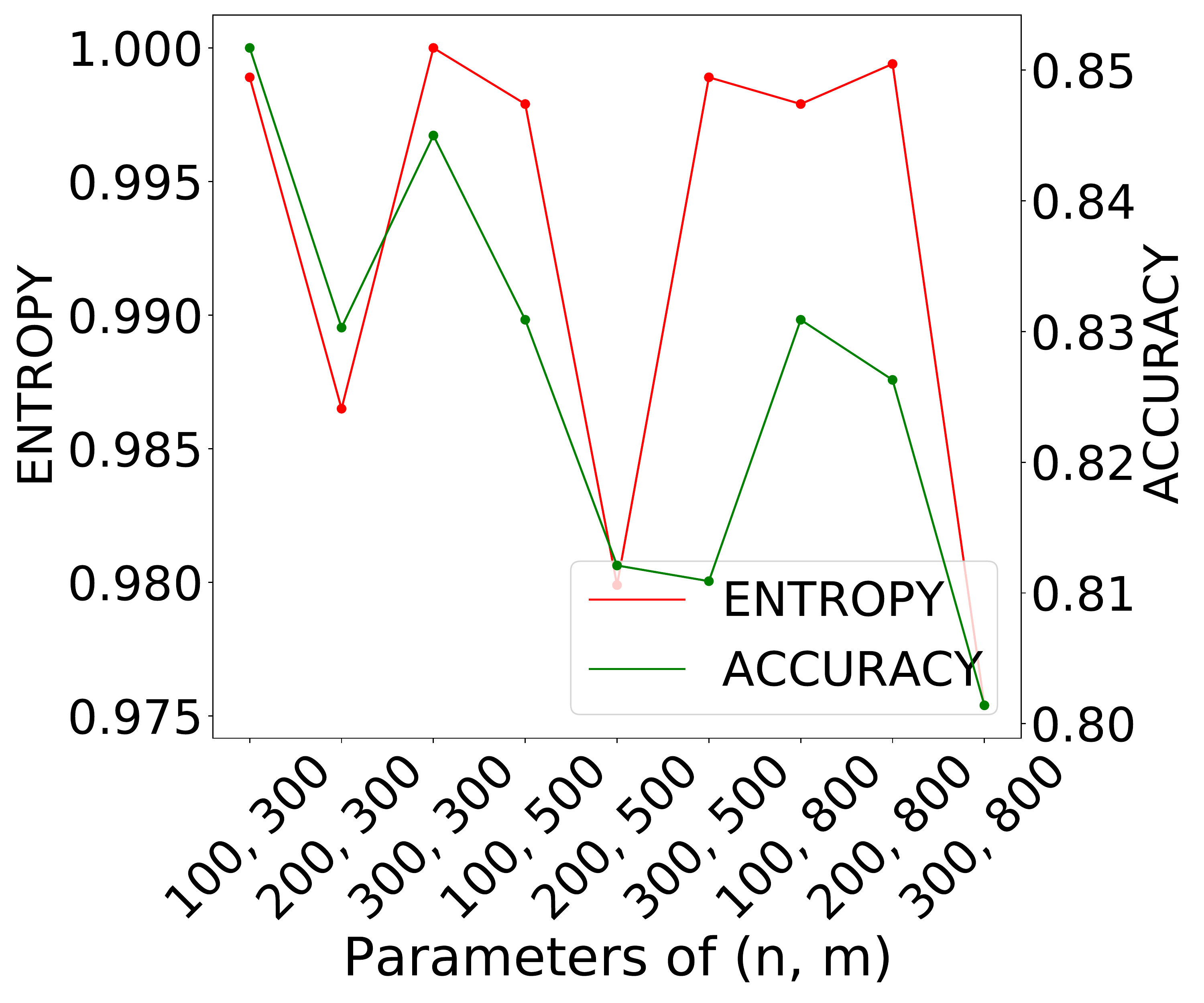}}
    \centerline{\small SST}
   \end{minipage}
   \begin{minipage}{.32\textwidth}
    \centerline
    {\includegraphics[width=\linewidth]{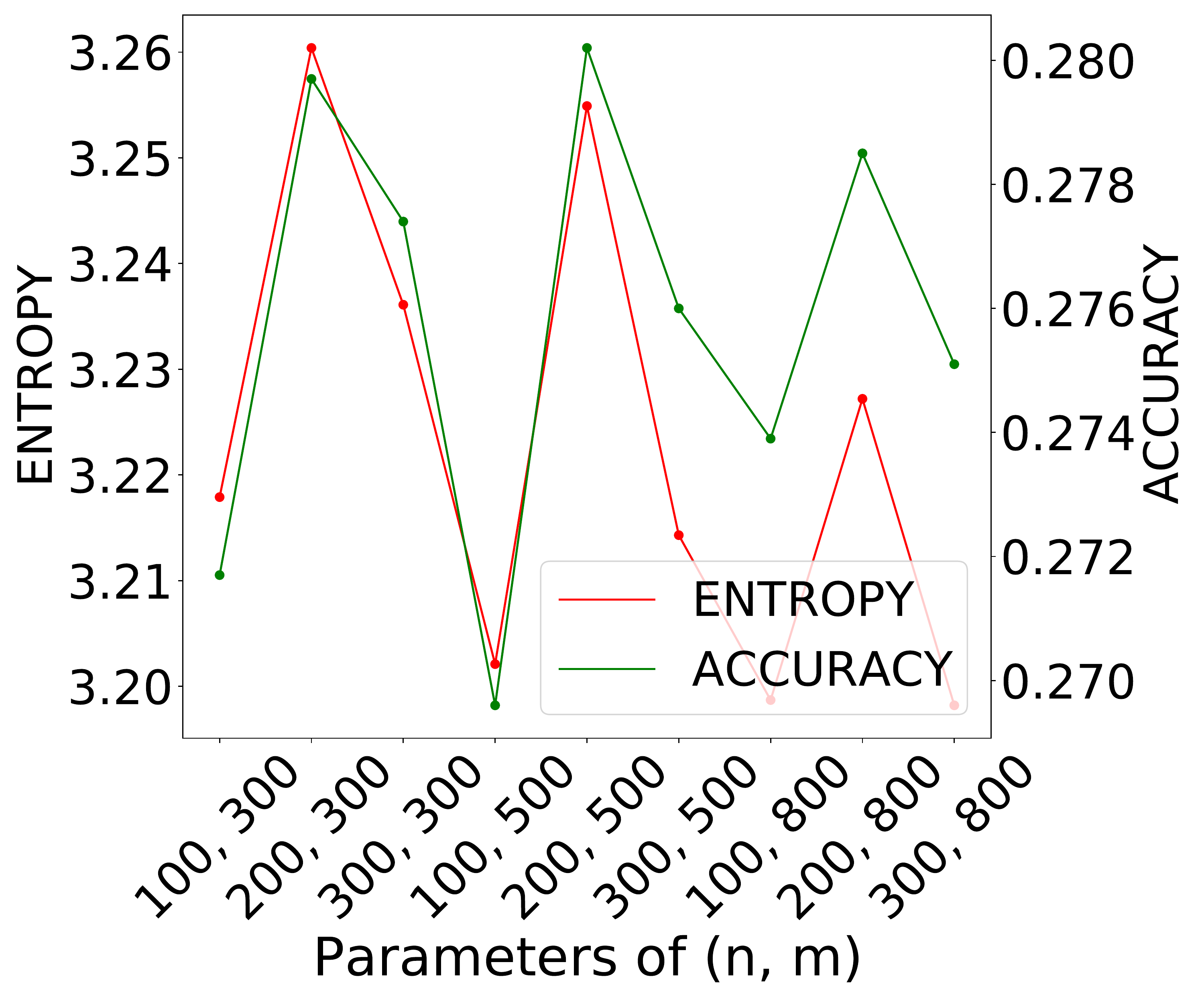}}
    \centerline{\small Emotion}
   \end{minipage}
  }
 
  \subfigure
  {
   \begin{minipage}{.32\textwidth}
    \centerline
    {\includegraphics[width=\linewidth]{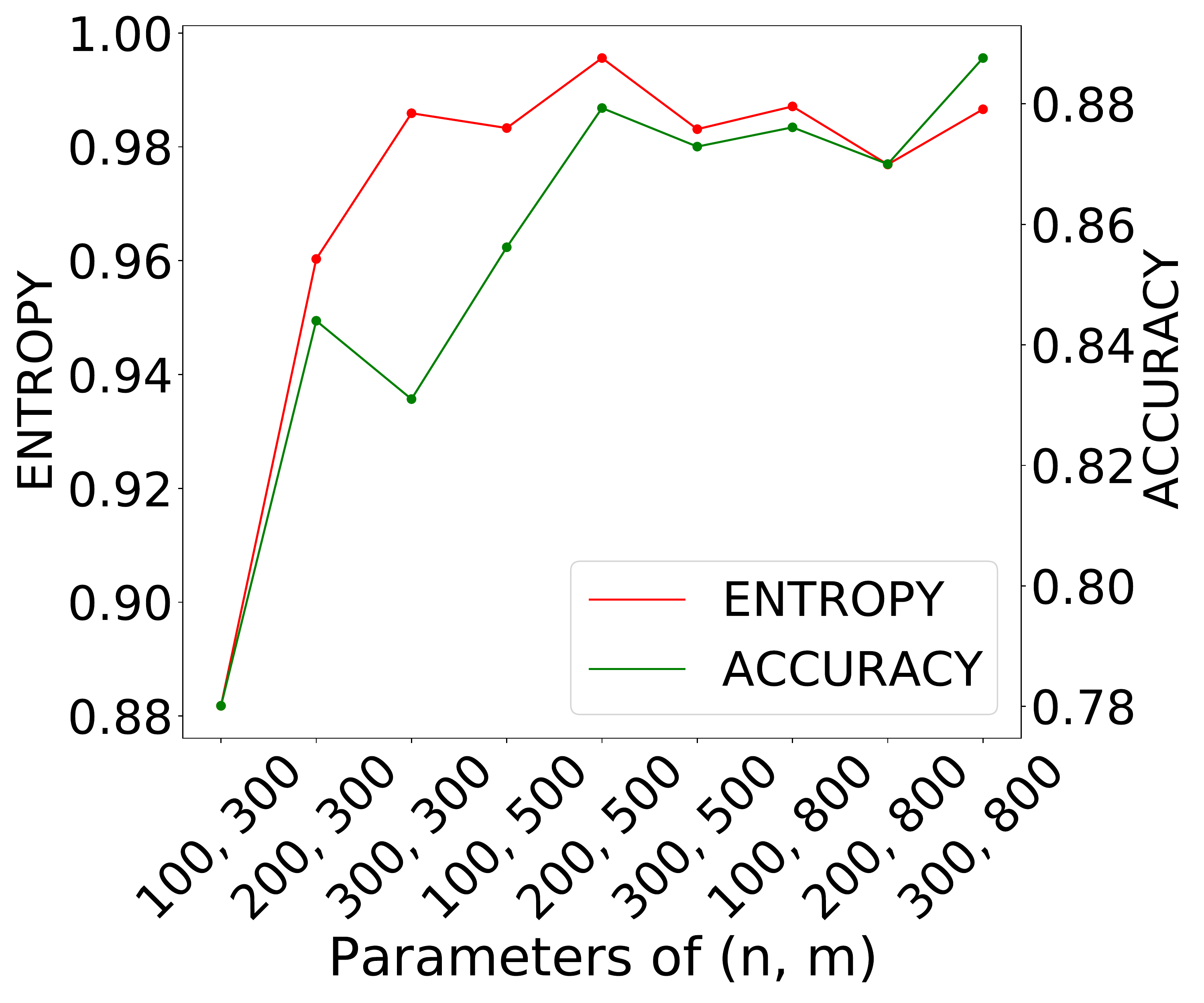}}
    \centerline{\small Yelp}
   \end{minipage}
   \begin{minipage}{.32\textwidth}
    \centerline
    {\includegraphics[width=\linewidth]{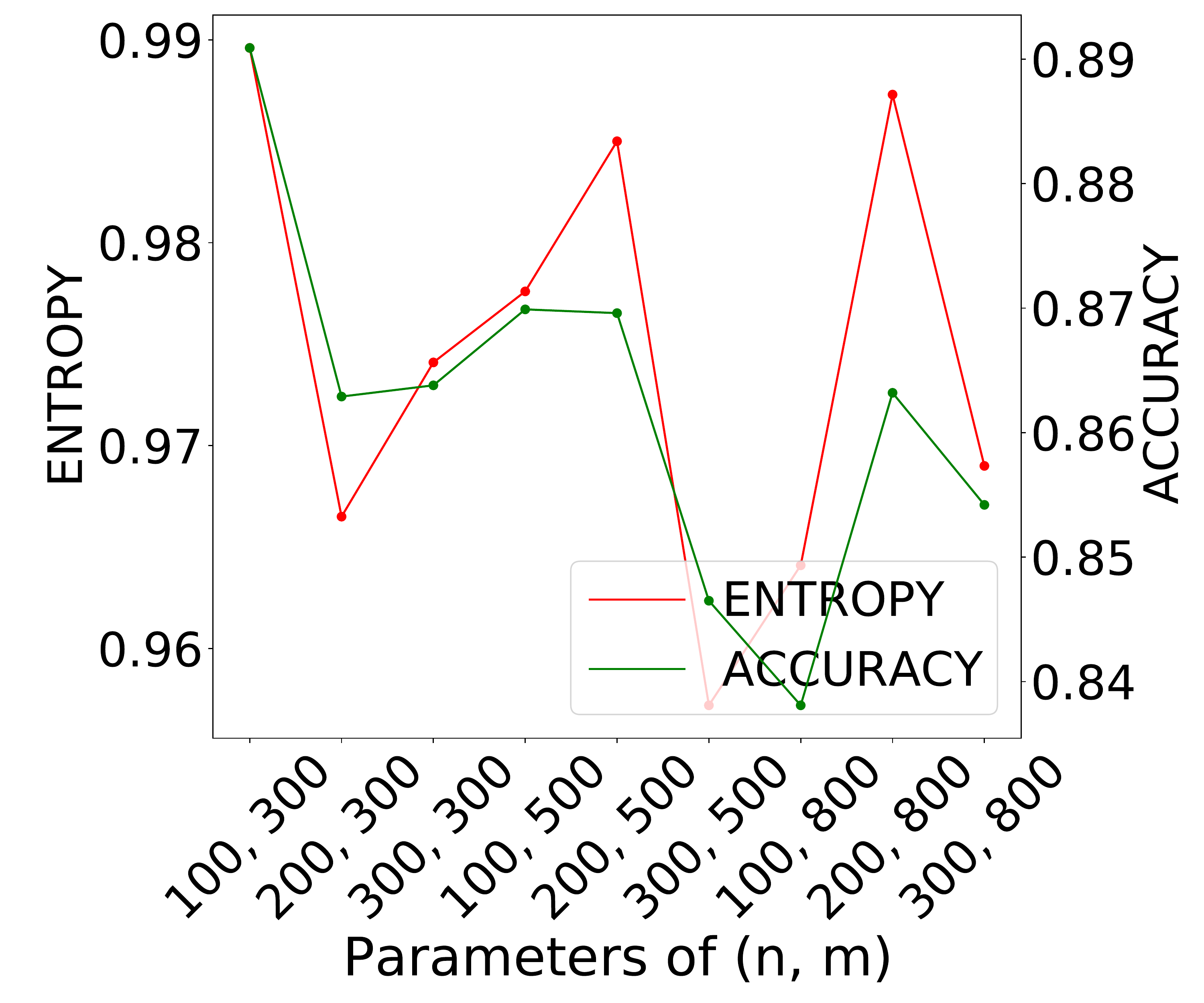}}
    \centerline{\small Amazon}
   \end{minipage}
   \begin{minipage}{.32\textwidth}
    \centerline
    {\includegraphics[width=\linewidth]{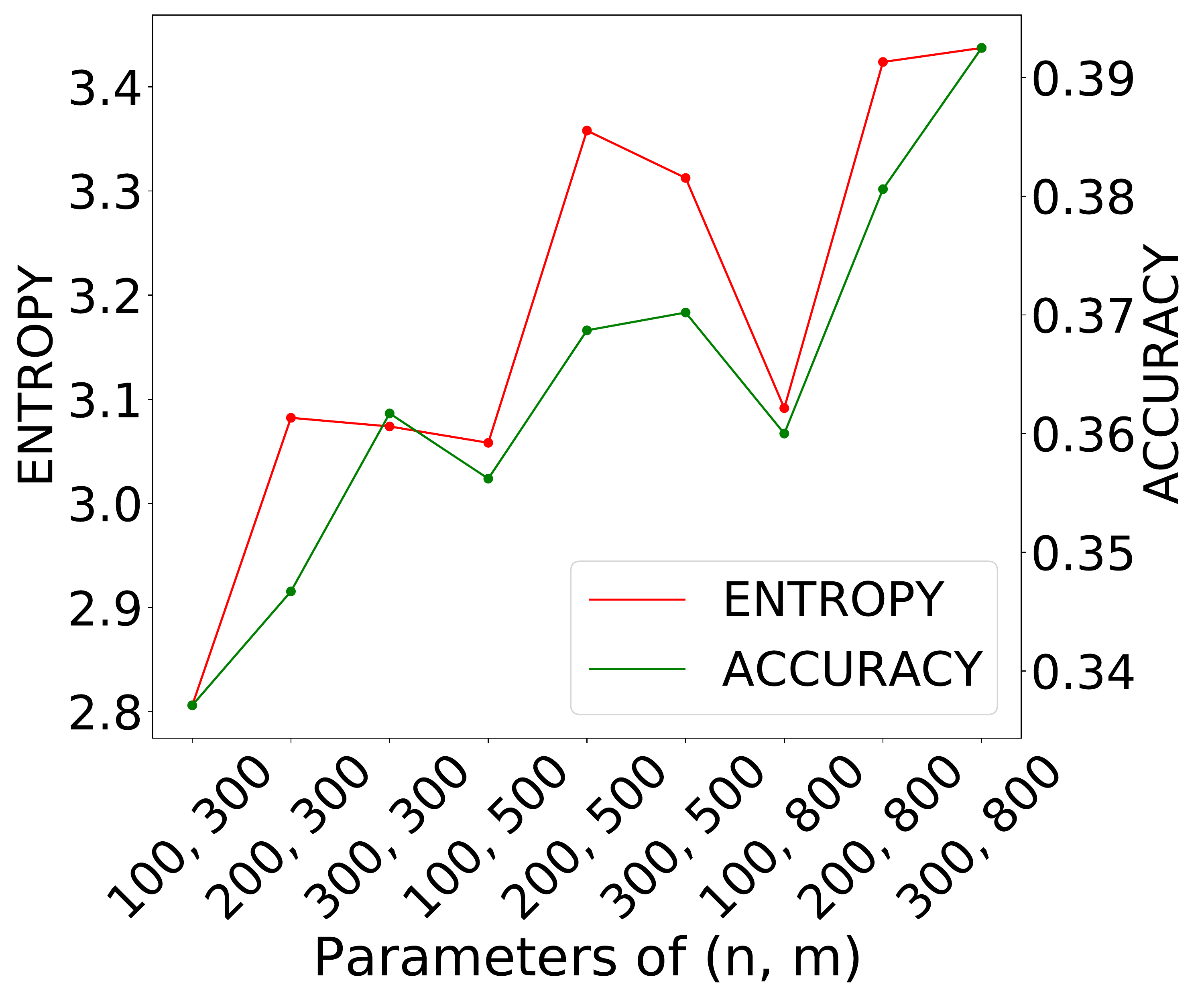}}
    \centerline{\small Situation}
   \end{minipage}
  }

 \caption{Relation between entropy and accuracy.}
 \label{entropy}
 \end{figure*}

In order to verify the relationship between entropy and classification accuracy, we compared the trends of entropy and predicate accuracy under different parameter settings. Figure \ref{entropy}  shows the relation between the entropy and accuracy in evaluation datasets. From Figure \ref{entropy} we can see that with different parameters, the trends of entropy and accuracy are often (but not perfectly) correlated. It shows that the empirical classification entropy on unlabeled data is an effective unsupervised metric to guide the selection of pseudo-labeled subset.

\end{document}